\documentclass[a4paper,fleqn]{cas-dc}

\usepackage[numbers]{natbib}
\usepackage{graphicx}
\usepackage{float}
\usepackage{todonotes}
\usepackage{adjustbox}
\usepackage[dvipsnames]{xcolor}

\PassOptionsToPackage{hyphens}{url}\usepackage{hyperref}
\usepackage[inline]{enumitem}

\def\tsc#1{\csdef{#1}{\textsc{\lowercase{#1}}\xspace}}
\tsc{WGM}
\tsc{QE}
\tsc{EP}
\tsc{PMS}
\tsc{BEC}
\tsc{DE}

\begin{document}\sloppy
\let\WriteBookmarks\relax
\def\floatpagepagefraction{1}
\def\textpagefraction{.001}
\shorttitle{Link Prediction of Scholarly Knowledge Graphs}
\shortauthors{M. Nayyeri et~al.}

\title [mode = title]{Trans4E: Link Prediction on Scholarly Knowledge Graphs} 





\author[1]{Mojtaba Nayyeri}[orcid=0000-0002-9177-0312]
\ead{nayyeri@cs.uni-bonn.de}

\author[1]{Gokce Muge Cil}[orcid=0000-0002-4645-0088]
\ead{s6gocill@uni-bonn.de}

\author[2]{Sahar Vahdati}[orcid=0000-0002-7171-169X]
\ead{vahdati@infai.org}

\author[4]{Francesco Osborne}[orcid=0000-0001-6557-3131]
\ead{francesco.osborne@open.ac.uk}

\author[1]{Mahfuzur Rahman}[orcid=0000-0003-0273-3112]
\ead{s6mrrahm@uni-bonn.de}

\author[5]{Simone Angioni}[orcid=0000-0002-6682-3419]
\ead{simone.angioni@unica.it}

\author[4]{Angelo Salatino}[orcid=0000-0002-4763-3943]
\cormark[1]
\ead{angelo.salatino@open.ac.uk}

\author[5]{Diego Reforgiato Recupero}[orcid=0000-0001-8646-6183]
\ead{diego.reforgiato@unica.it}

\author[1]{Nadezhda Vassilyeva}[orcid=0000-0000-0000-0000]
\ead{vassilyeva@cs.uni-bonn.de}

\author[4]{Enrico Motta}[orcid=0000-0003-0015-1952]
\ead{enrico.motta@open.ac.uk}

\author[1,3]{Jens Lehmann}[orcid=0000-0001-9108-4278]
\ead{jens.lehmann@cs.uni-bonn.de}

\address[1]{SDA Research Group, University of Bonn (Germany)}
\address[2]{Institute for Applied Informatics (InfAI)}
\address[3]{Fraunhofer IAIS, Dresden (Germany)}
\address[4]{Knowledge Media Institute, The Open University, Milton Keynes (UK)}
\address[5]{Department of Mathematics and Computer Science, University of Cagliari (Italy)}

\cortext[cor1]{Corresponding author}


\begin{abstract}
The incompleteness of Knowledge Graphs (KGs) is a crucial issue affecting the quality of AI-based services. 
In the scholarly domain, KGs describing research publications typically lack important information, hindering our ability to analyse and predict research dynamics.
In recent years, link prediction approaches based on Knowledge Graph Embedding models became the first aid for this issue.
In this work, we present Trans4E, a novel embedding model that is particularly fit for KGs which include N to M relations with N$\gg$M. This is typical for KGs that categorize a large number of entities (e.g., research articles, patents, persons) according to a relatively small set of categories. Trans4E was applied on two large-scale knowledge graphs, the Academia/Industry DynAmics (AIDA) and Microsoft Academic Graph (MAG), for completing the information about Fields of Study (e.g., 'neural networks', 'machine learning', 'artificial intelligence'), and affiliation types (e.g., 'education', 'company', 'government'), improving the scope and accuracy of the resulting data.
We evaluated our approach against alternative solutions on AIDA, MAG, and four other benchmarks (FB15k, FB15k-237, WN18, and WN18RR). Trans4E outperforms the other models when using low embedding dimensions and obtains competitive results in high dimensions. 

\end{abstract}

%
%

\begin{keywords}
Scholarly Knowledge Graph \sep
Knowledge Graph Embedding \sep
Scholarly Communication \sep
Science Graph \sep
Metaresearch Queries \sep
Link Prediction \sep
Research of Research
\end{keywords}

\maketitle

\section{Introduction}

The technology of Knowledge Graphs (KGs) empowered by graph-based knowledge representation brought an evolutionary change in a range of AI tasks.
As a consequence, many application domains in science, industry, and different enterprises use KGs for data management. 
However, a challenge with KGs is that, despite the presence of millions of triples, capturing complete knowledge from the real world is almost impossible, even for specific application domains.
Therefore, KGs usually remain incomplete. 

Scientific research is one of the major domains for the application of KGs. 
In the last years, we saw the emergence of several KGs describing research outputs, such as  
Microsoft Academic Graph\footnote{Microsoft Academic Graph - \url{http://aka.ms/microsoft-academic}}~\cite{wang2020microsoft}, 
Scopus\footnote{Scopus - \url{https://www.scopus.com/}}, 
Semantic Scholar\footnote{Semantic Scholar - \url{https://www.semanticscholar.org/}}, Aminer~\cite{zhang2018name},
Core~\cite{knoth2012core}, 
OpenCitations~\cite{peroni2020opencitations}, Dimensions\footnote{Dimensions - \url{https://www.dimensions.ai/}}, 
Open Research Knowledge Graph\footnote{ORKG - \url{https://www.orkg.org/orkg/}}~\cite{jaradeh2019open}, and others.
These solutions are crucial for performing large-scale bibliometric studies, 
informing funding agencies and research policymakers, supporting a variety of intelligent systems for querying the scientific literature, identifying research topics, suggesting relevant articles and experts, detecting research trends, and so on. 
Their usefulness and, consequently, our ability to assess research dynamics, are however crucially limited by their incompleteness. 
Even basic metadata such as affiliations, organization types, references, research topics, and conferences are often missing, noisy, or not properly disambiguated. Therefore, apparently simple tasks such as identifying the affiliation and the country of origin of a publication still require a large amount of manual data cleaning~\cite{mannocci2019geographical}.

Traditionally, data integration methods have been applied to solve data incompleteness in the context of databases and repositories. However, when completing and refining large KGs, it is crucial to adopt scalable and automatic approaches.
Among the many possible graph completion methods, Knowledge Graph Embedding (KGE) models have recently gained a lot of attention.
KGEs learn representations of graph nodes and edges with the goal of predicting links between existing entities. 
Embedding models have been in practical use for various types of KGs in different domains, including digital libraries~\cite{yao2017incorporating}, biomedical~\cite{li2019prtransh}, and social media~\cite{stanovsky2017recognizing}. 


%
%



However, the specific characteristics of scholarly KGs poses important challenges for link prediction methods based on KGE models~\cite{bordes2013translating, sun2018rotate, trouillon2016complex, quate2019zhang, wang2020attributed,tran2019exploring,nayyeri2020embedding}. 
One crucial aspect is the presence of several N to M relations with N$\gg$M.\ Given a triple ($h,r,t$), this situation arises when the cardinality of the entities in the head position ($h$) for a certain relation ($r$) is much higher than the one of the entities in the tail position ($t$). 
This is the case for most scholarly knowledge graphs~\cite{peroni2020opencitations,wang2020microsoft,ammar2018construction,zhang2018name,core2011} that usually categorize millions of documents (e.g., papers, patents) 
according to a relatively small set of categories (e.g., topics, affiliation kinds, countries, chemical compounds).
Current KGE models lack the ability to handle effectively these kinds of relations since they are unable to assign to each entity a well distinct embedding vector in a low dimensional space. As a result, link prediction and node classification techniques that exploit these embeddings tend to perform poorly. 


To address this problem, we propose Trans4E, a new embedding model specifically designed to support link prediction for KGs which present N to M relations with N$\gg$M. 
Specifically, Trans4E tackles the issue by providing a larger number of possible vectors ($8^d-1$, where $d$ is the embedding dimension) to be assigned to entities involved in N to M relations. Trans4E enables the generation of a well distinct vector for each entity even when using small embedding dimensions.

The motivating scenario for this work was supplied by the Academia/Industry DynAmics (AIDA)\footnote{AIDA - \url{http://aida.kmi.open.ac.uk}} Knowledge Graph~\cite{angioni2020integrating}, a resource that  was designed for studying the relationship between academia and industry and for supporting systems for predicting research dynamics. The current version of AIDA integrates the metadata about 21M publications from Microsoft Academic Graph (MAG) and 8M patents from Dimensions in the field of Computer Science. 
In this resource, documents are categorized according to their research topics drawn from the Computer Science Ontology (CSO)\footnote{CSO - \url{https://cso.kmi.open.ac.uk/}}~\cite{salatino2020computer} and classified with their authors' affiliation types on the Global Research Identifier Database (GRID)\footnote{GRID - \url{https://www.grid.ac/}} (e.g., `Education', `Company', `Government', `Nonprofit'). 
This solution enables analysing the evolution of research topics across academia, industry, government institutions, and other organizations. For instance, it allows us to detect that a specific topic, originally introduced by academia, has been recently adopted by industry. 
It can also support systems for predicting the impact of specific research efforts on the industrial sector~\cite{salatino2020research}
and the evolution of technologies~\cite{osborne2017}.
Nevertheless, only 5.1M out of the 21M articles could be mapped to a GRID and characterized according to their affiliation type. Therefore, more than 75\% of the publications are missing this critical information, significantly reducing the scope and accuracy of the resulting analytics. 
In order to show that our approach can be applied to fields with very different characteristics, we also use it to complete the Fields of Study, which is a collection of terms from multiple disciplines utilized to index the articles in MAG. Indeed, the completeness of the set of terms associated with a paper varies a lot and depends on the quality and style of the abstract, which in turn is often parsed from online PDFs, leading to mistakes and missing content. This in turn hinders our ability to understand the research concepts associated with the paper and to obtain comprehensive analytics.
Completing the affiliation types and the Fields of Study is crucial for improving the overall quality of these knowledge graphs and a very good practical use case for link prediction.



We evaluated Trans4E against several alternative models (TransE, RotatE, QuatE, ComplEx) on the task of link prediction on AIDA, MAG, and four other well-known benchmarks (FB15K, FB15k-237, WN18, and WN18RR).


The experiments showed that Trans4E outperforms the other approaches in the case of N to M relations with N $\gg$ M and yields very competitive results in all the other cases, in particular when using low embedding dimensions.
The ability to solve the N $\gg$ M issue and to perform well even when adopting small embedding dimensions makes Trans4E particularly apt for handling large scale knowledge graphs that describe millions of entities of the same type (e.g., documents, persons).

In summary, the contributions of our work are the following:
\begin{itemize}
\item We propose Trans4E, a new embedding model specifically designed to provide link prediction for large-scale KGs presenting N to M relations with N $\gg$ M.

\item We apply Trans4E on a real word scenario that involves completing affiliation types and Fields of Study (N $\gg$ M relations) in AIDA and MAG.

\item We further evaluate our approach on four well-known benchmarks (FB15k, FB15k-237, WN18, and WN18RR), showing that Trans4E yields competitive performances in several configurations. 
\end{itemize}


%

The rest of the paper is organised as follows. In Section 2, we review the literature on current embedding models for data completion and scholarly knowledge graphs. In Section 3, we present a motivating scenario involving the completion of the AIDA knowledge graph. In Section 4, we describe Trans4E.  Section 5 reports the evaluation of the model versus alternative solutions. Finally, in Section 6 we summarise the main conclusions and outline future directions of research.

%

\section{Background and Related Work}
\label{sec:relatedwork}

In this section, we will first review the graph embedding models and their application to link prediction. Then we will discuss the current generation of scholarly knowledge graphs that can benefit from these solutions.

\subsection{Knowledge Graph Embedding Models}
In this section, we introduce the definitions required to understand our approach. 

\textbf{Embedding Vectors.}
Let the knowledge graph be $KG = (\mathcal{E},\mathcal{R}, \mathcal{T})$, where  $\mathcal{E}$ is the set of entities (nodes) in the graph, $\mathcal{R}$ is the set of all relations (edges), and $\mathcal{T}$ is the set of all triples in the graph in the form of $(h,r,t)$, e.g.,~\textit{(Berlin, CapitalOf, Germany)}.    
KGE models are applied to KGs for link prediction by measuring the degree of correctness of a triple. 
To do so, a KGE model aims at mapping each entity and relation of the graph into a vector space (shown as ($\mathbf{h}, \mathbf{r}, \mathbf{t}$), $\mathbf{h,r,t} \in \mathbb{R}^d$), where \textit{d} is the embedding dimension of each vector.
By $h_i$, we refer to the $i$-th element of the vector $\mathbf{h}$ where $i$ ranges in $\{1,\ldots ,d\}$. 
The vector representation of the entities and the relations in a KG are the actual embeddings. 

\textbf{Score Function.} 
Using this representation, the plausibility of the triples is then assessed by the scoring function $f(\mathbf{h},\mathbf{r},\mathbf{t})$ of the applied KGE model.
If a triple is more plausible, its score should be higher. For example, $f(\textit{Berlin, CapitalOf, Germany})$ should be higher than $f(\textit{Berlin, CapitalOf, France})$. 

\textbf{Negative Sampling.} 
Typical machine learning approaches are trained on both positive and negative samples. 
However, all the triples present in KGs are considered true, and this necessitates the injection of negative samples in the training of the KGEs. In this work we use Adversarial negative sampling (\textit{adv}) for this purpose. This technique generates a set of negative samples from a triple ($h,r,t$) by using a probabilistic algorithm to replace $h$ or $t$ with a random entity ($h'$ or $t'$) existing in $\mathcal{E}$.

\textbf{Loss Function.}
Since at the beginning of the learning process, the embedding vectors are initialized with random values, the scores of the triples for positive and negative samples are also random.
Optimization of a loss function $\mathcal{L}$ is utilized to adjust the embeddings in such a way that positive samples get higher scores than the negatives ones.  
Stochastic Gradient Descent (SGD) method is commonly used for optimising the loss function.

\textbf{N to M Relations.}
As mentioned above, given a relation $r$, the representation of facts in triple form is $(h,r,t)$.
Depending on the type of a relation and its meaning, for a fixed head (say $h_1$), there are at most M possible tails connected to the head, i.e.~$\{(h_1, r, t_1), (h_1, r, t_2)\}, \ldots, (h_1, r, t_M)$. Similarly, for a fixed tail, (say $t_1$), there are at most N possible head entity, i.e.~$\{(h_1, r, t_1),(h_2, r, t_1), \ldots, (h_N, r, t_1)\}$. 
There are four cases that may arise for a relations which connects a different number of heads and tails: a) both N and M are small, b) both M and N are large, c) N is small and M is large, and d) N is large and M is small. 
The latter is the focus of this paper.
For example, in the AIDA knowledge graph the ``hasType'' relations connects a very large number of head entities (5.1M articles) to only 8 tail entities (the GRID types).

\subsection{Review of State-of-the-art KGEs}

Here we summarize some of the most used existing models focusing in particular on their scoring function. 
\paragraph{TransE}\cite{bordes2013translating} is one of the early embedding models and is well known for its outstanding performance and simplicity. 
It is a solid baseline that can still outperform many of the most recent and complex KGEs \cite{henck2019}.
The idea of the TransE model is to enforce embedding of entities and relations in a positive triple ($h,r,t$) to satisfy the following equality:
 \begin{align}
     \textbf{h} + \textbf{r} \approx \textbf{t}
 \label{eq:tran}     
 \end{align}
where \textbf{h}, \textbf{r} and \textbf{t} are the embedding vectors of head, relation, and tail, respectively. TransE model defines the following scoring function:
\begin{align}\label{eq:TransE_score}
    f_r(h,t) = -\| \textbf{h} + \textbf{r} - \textbf{t} \|
\end{align}

\paragraph{RotatE}\cite{sun2018rotate} is a model designed to transform the head entity to the tail entity by using the relation rotation. 
This model embeds entities and relations in complex space. 
If we constrain the norm of entity vectors, this model would be reduced to TransE.
The scoring function of RotatE is 
\begin{equation}
    f_r(h,t) = -\| \textbf{h} \circ \textbf{r} - \textbf{t} \|
\end{equation}
in which $\circ$ is the element-wise product.
Rotate is one of the recent state-of-the-art models which is leading the accuracy competition among KGEs \cite{sun2018rotate}.  


\paragraph{ComplEx}\cite{trouillon2016complex}
 is a semantic matching model, which assesses the plausibility of facts by considering the similarity of their latent representations.
In other words, it is assumed that similar entities have common characteristics, i.e.\ are connected through similar relationships
\cite{Nickel2015ReviewRelationalMLforKG,wang2017knowledge}.
In ComplEx the entities are embedded in the complex space.
The score function of ComplEx is given as follows:
$$
f(h,t)=
\Re(\mathbf{h}^{T}\,\text{diag}(\mathbf{r})\,\bar{\mathbf{t}})
$$
in which $\bar{\mathbf{t}}$ is the conjugate of the vector $\mathbf{t}$ and $\Re$ returns the real part of the complex number. 

\paragraph{QuatE}\cite{quate2019zhang} 
model relations in the quaternion
space.
Similarly to RotatE, QuatE represents a relation as a rotation. 
However, a rotation in quaternion space is more expressive than a rotation in complex space. 
A product of two quaternions $ Q_{1}\otimes Q_{2} $ is equivalent to first scaling $ Q_{1} $ by magnitude $ |Q_{2}| $ and then rotating it in four dimensions.
QuatE finds a mapping $ \mathcal{E} \rightarrow  \mathbb{H}^{d} $, where an entity \textit{h} is represented by a quaternion vector $ \mathbf{h} = a_{h} + b_{h}\mathbf{i} + c_{h}\mathbf{j} + d_{h}\mathbf{k} $, with $ a_{h}, b_{h}, c_{h}, d_{h} \in \mathbb{R}^{d}$.

The scoring function is computed as follows:
\begin{equation}\label{eq:quate}
\phi(h,r,t) = \mathbf{h'\cdot t} = \langle a'_{h},a_{t}  \rangle  + \langle b'_{h}, b_{t} \rangle  + \langle  c'_{h}, c_{t}\rangle + \langle d'_{h}, d_{t} \rangle
\end{equation}
where $ \langle \cdot, \cdot\ \rangle $ is the inner product. $ \mathbf{h'} $ is computed by first normalizing the relation embedding \textbf{r} = $ p_{r} + q_{r}\mathbf{i} + u_{r}\mathbf{j}+v_{r}\mathbf{k} $ to a unit quaternion: 
\begin{equation}
\mathbf{r}^{(n)} = \dfrac{r}{|r|} = \dfrac{p_{r} + q_{r}\mathbf{i} + u_{r}\mathbf{j}+v_{r}\mathbf{k}}{\sqrt{ p_{r}^{2} + q_{r}^{2} + u_{r}^{2}+v_{r}^{2} }}
\end{equation}
and then computing the Hamiltonian product between $ \mathbf{r}^{(n)} $ and \textbf{h} = $ a_{h} + b_{h}\mathbf{i} + c_{h}\mathbf{j} + d_{h}\mathbf{k}  $:

\begin{equation}
\begin{aligned}
\mathbf{h}^{\prime}=\mathbf{h} \otimes \mathbf{r}^{(n)} &:=\left(a_{h} \circ p-b_{h} \circ q-c_{h} \circ u-d_{h} \circ v\right) \\
&+\left(a_{h} \circ q+b_{h} \circ p+c_{h} \circ v-d_{h} \circ u\right) \mathbf{i} \\
&+\left(a_{h} \circ u-b_{h} \circ v+c_{h} \circ p+d_{h} \circ q\right) \mathbf{j} \\
&+\left(a_{h} \circ v+b_{h} \circ u-c_{h} \circ q+d_{h} \circ p\right) \mathbf{k}
\end{aligned}
\end{equation}

\subsection{Further Related Work}
Beside KGE models such as QuatE, ComplEx, TransE, and RotatE, there are several related approaches based on neural networks for KG completion.
Here we cover the most relevant ones with specific focus on link prediction.

Few-Shot Learning (FSL) takes advantage of prior knowledge for efficiently learning from a limited number of examples \cite{wang2020generalizing,yin2020meta}. 
Some FSL methods focus on graph meta-learning \cite{bendre2020learning,zhang2020few,bose2019meta}, which provide fast adaption to the newly imported data.
Such techniques are not reported to be suitable for large scale KGs and are mainly used on image datasets.
Some other FSL methods focus on link prediction on graphs. For instance, authors in \cite{chen2019meta} use meta information for learning the most important and relevant knowledge with high retrieval speed. 
This model is similar to TransE and therefore suffers from the same limitation regarding N to M relation types with N $\gg$ M. 
In addition, FSL is most useful in scenarios with few examples. This is not the case of large-scale scholarly knowledge graph that usually can produce a large number of examples.  


Transfer Learning (TL) aims at reusing the knowledge gained while solving a specific problem for solving a different but related one~\cite{zhuang2020comprehensive}.
This strategy enables to address new learning tasks without extensive re-training~\cite{gong2019graphonomy,paliwal2019regal}. 
A family of methods labelled Graph Transfer Learning (GTLs) are specifically designed to work on graph-structured data~\cite{lee2017transfer}.
However, they are not directly applicable to the task discussed in this paper since they mostly focus on similarity between entities rather than link prediction.

Graph Neural Networks (GNNs) are often used for link prediction~\cite{arora2020survey,wu2020comprehensive}.
These methods compute the state of embeddings for a node according to the local neighborhood~\cite{zhou2018graph,wang2019robust,wang2019knowledge}.
The embedding of a node is then created as a $d$-dimension vector and produces output information such as the node label.
However, the high computation costs of GNNs make them unsuitable for large-scale knowledge graphs. 

Several approaches for link prediction based on Deep Neural Networks, such as KBAT~\cite{nathani2019learning} and  CapsE~\cite{vu2019capsule}, reported very high performance, but were not consistent across different benchmarks. An analysis by Su at al.~\cite{sun2019re} showed that this behaviour was due to an inappropriate evaluation protocol, and the performance of these models dropped after fixing the relevant biases.
According to them, instead, shallow KGE models (e.g.,  TransE, RotatE, ComplEx, QuatE) are able to perform consistently across several evaluation protocols~\cite{sun2019re}.


Another interesting family of approaches regards community embeddings, which can optimize node embedding by using a community-aware high-order proximity~\cite{cavallari2019embedding}. For instance, vGraph~\cite{sun2019vgraph} is a probabilistic generative model that learns community membership and node representation collaboratively. 
ComE+~\cite{cavallari2019embedding} is another approach for community embedding which can tackle the situation in which the number of communities is unknown. 
However, these methods focus on node and community embeddings based on intra-group connections in terms of clustering and node classification tasks rather than link prediction.

Details of more relevant works are discussed in several surveys~\cite{bonatti2019knowledge,chen2020review}. 
A comprehensive review of recent approaches for knowledge graph representation is available in a survey paper~\cite{ji2020survey}, where KG embedding models are discussed in terms of representation space, scoring function, encoding models, and auxiliary information. 
The survey also discusses a broad range of reasoning methods such as Random Walk inference, Deep Reinforcement learning for multi-hop reasoning, and rule-based reasoning which however mainly focus on path-related reasoning rather than link prediction.







\subsection{Scholarly Knowledge Graphs}

 In the last years, we saw the emergence of several knowledge graphs describing research publications.
 Traditionally, they either focus on the metadata of the articles, such as titles, abstracts, authors, organizations, or, more rarely, they offer a machine-readable representation of the  knowledge contained therein. 

A good example of the first category is Microsoft Academic Graph (MAG)~\cite{wang2020microsoft}, which 
is a heterogeneous knowledge graph containing the metadata of more than 242M scientific publications, including  citations, authors, institutions, journals, conferences, and Fields of Study. 
Similarly, the Semantic Scholar Open Research Corpus\footnote{ORC - \url{http://s2-public-api-prod.us-west-2.elasticbeanstalk.com/corpus/}}~\cite{ammar2018construction} is a dataset of about 185M publications  released by Semantic Scholar, an academic search engine provided by the Allen Institute for Artificial Intelligence.
The OpenCitations Corpus~\cite{peroni2020opencitations} is released by OpenCitations, an independent infrastructure organization for open scholarship dedicated to the publication of open bibliographic and citation data with semantic technologies. The current version includes 55M  publications and 655M citations. 
Scopus is a well-known dataset curated by Elsevier, which includes about 70M publications and is often used by governments and funding bodies to compute performance metrics.
The AMiner Graph~\cite{zhang2018name} is a corpus of more than 200M publications generated and used by the AMiner system\footnote{AMiner - \url{https://www.aminer.cn/}}. AMiner is a free online academic search and mining system that also extracts researchers’ profiles from the Web and integrates them in the metadata.
The Open Academic Graph (OAG)\footnote{OAG - \url{https://www.openacademic.ai/oag/}} is a large knowledge graph integrating Microsoft Academic Graph and AMiner Graph. The current version contains 208M papers from MAG and 172M from AMiner.
Core~\cite{core2011}\footnote{CORE - \url{https://core.ac.uk/}} is a  repository that integrates 24M open access research outputs from repositories and journals worldwide. The Dimensions Corpus is a dataset produced by Digital Science which integrates and interlinks 109M research publications, 5.3M grants, and 40M patents. 
 
All these resources suffer from different degrees of data incompleteness. For instance, it is still challenging to identify and disambiguate affiliations, which also hinders the ability to categorize the articles according to their affiliation types or countries~\cite{mannocci2019geographical}. Similarly, references are usually incomplete, and the citation count of the same paper tends to vary dramatically on different datasets ~\cite{peroni2020opencitations}.

A second category of knowledge graphs focuses instead on representing the content of scientific publications. This challenging objective was traditionally pursued by the semantic web community, e.g., by creating bibliographic repositories in the Linked Data Cloud ~\cite{nuzzolese2016semantic}, generating knowledge bases of biological data~\cite{belleau2008bio2rdf}, encouraging  the Semantic Publishing paradigm~\cite{shotton2009semantic}, formalising research workflows~\cite{wolstencroft2013taverna}, implementing systems for managing nano-publications~\cite{groth2010ana,kuhn2016decentralized} and  micropublications~\cite{schneider2014using}, developing a variety of ontologies to describe scholarly data, e.g., SWRC\footnote{SWRC - \url{http://ontoware.org/swrc}}, BIBO\footnote{BIBO - \url{http://bibliontology.com}}, BiDO\footnote{BiDO - \url{http://purl.org/spar/bido}}, SPAR~\cite{peroni2018spar}\footnote{SPAR - \url{http://www.sparontologies.net/}}, CSO\footnote{CSO - \url{http://http://cso.kmi.open.ac.uk}}~\cite{salatino2018computer}. A recent example is the Open Research Knowledge Graph (ORKG)~\cite{jaradeh2019open}\footnote{ORKG - \url{https://www.orkg.org/orkg/}}, which aims to describe research papers in a structured manner to make them easier to find and compare. Similarly, the Artificial Intelligence Knowledge Graph (AI-KG)~\cite{dessi2020ai}\footnote{AI-KG - \url{http://scholkg.kmi.open.ac.uk/}} describes 1.2M statements extracted from 333K research publications in the field of AI.
Since extracting the scientific knowledge from research articles is still a very challenging task,  these resources tend also to suffer from data incompleteness. Therefore, it is crucial to develop new models that could tackle this issue and improve the quality of these KGs.




 %

\section{Motivating Scenario: AIDA Knowledge Graph}
\label{sec:kg}
%
%

\subsection{Incompleteness in AIDA}


Academia, industry, public institutions, and non-profit organizations collaborate in the crucial effort of advancing scientific knowledge. Analysing the knowledge flow between them, assessing the best policies to harmonise their efforts, and detecting how they address emerging research areas is a critical task for researchers, funding bodies, and companies in the space of innovation.
However, today scholarly KGs ~\cite{peroni2020opencitations,wang2020microsoft,ammar2018construction,zhang2018name,core2011} do not support well this task since they typically lack a high-quality characterization of the research topics, affiliation types, and industrial sectors. Therefore, we recently introduced the Academia/Industry DynAmics (AIDA) Knowledge Graph~\cite{angioni2020integrating}, 
which includes more than one billion triples and describes 20M publications from Microsoft Academic Graph (MAG)\footnote{MAG - \url{https://academic.microsoft.com/}}~\cite{wang2020microsoft} and 8M patents from Dimensions\footnote{Dimensions - \url{https://www.dimensions.ai/}} according to the 14K research topics from the Computer Science Ontology (CSO)\footnote{CSO - \url{http://cso.kmi.open.ac.uk/}}~\cite{salatino2020computer}. 
In addition, 5.1M publications and 5.6M patents that were associated with IDs from the Global Research Identifier Database (GRID)\footnote{GRID - \url{https://www.grid.ac/}} in the original data were also classified according to  the type of the author's affiliations and 66 industrial sectors (e.g., automotive, financial, energy, electronics) drawn from the Industrial Sectors ontology (INDUSO)\footnote{INDUSO - \url{http://aida.kmi.open.ac.uk/downloads/induso.ttl}}. 
The mapping with MAG enables to characterize all articles according to the relevant scholarly entities in the MAG~\cite{wang2020microsoft,farber2019microsoft}, including \textit{authors}, \textit{conferences}, \textit{journals},  \textit{references} (the full citation network), and the \textit{Fields of Study}. In the following, we will refer to the combination of the two knowledge graph as \textit{AIDA+MAG}. 

AIDA is available at \textbf{\url{http://aida.kmi.open.ac.uk}} and can be downloaded as a dump or  queried via a Virtuoso triplestore (\textbf{\url{http://aida.kmi.open.ac.uk/sparql/}}).  The AIDA ontology builds on SKOS, CSO, and INDUSO and it is available at \textbf{\url{http://aida.kmi.open.ac.uk/ontology}}.


%


\begin{table}[]
\caption{AIDA - Number of documents associated with the main GRID types.}
    \label{tab:overview}
\begin{tabular}{l|c|c}
\toprule
\multicolumn{1}{l|}{} & \multicolumn{1}{l|}{ \textbf{Papers}} & \multicolumn{1}{l}{ \textbf{Patents}} \\ 
\midrule
 Education             &  3,969,097                            &  169,884                              \\
 Company               &  954,143                              &  5,335,836                            \\
 Government            &  185,633                              &  54,396                               \\
 Facility              &  169,234                              &  66,605                               \\
 Nonprofit             &  61,129                               &  38,959                               \\
 Healthcare            &  28,362                               &  28180                                \\
 Other                 &  25,028                               &  16,631                               \\ \midrule
 \textbf{Typed (GRID)}          &  5,133,171                            &  5,639,252                            \\
 \textbf{Total}                 &  20,850,710                           &  7,940,034                            \\ \bottomrule
\end{tabular}
\end{table}




Table~\ref{tab:overview} shows the number of publications and patents associated with the main categories from GRID. A document can be associated with multiples types according to the affiliations of the creators. Academic institutions (`education') are responsible for the majority of research publications (77.5\%), while companies contribute to 19.8\% of them. When considering patents, the picture is very different: 94.6\% of them are from companies and only 3.0\% are typed as 'education'. 

\begin{figure*}[]
\centering
\minipage{0.495\textwidth}
  \includegraphics[width=\linewidth]{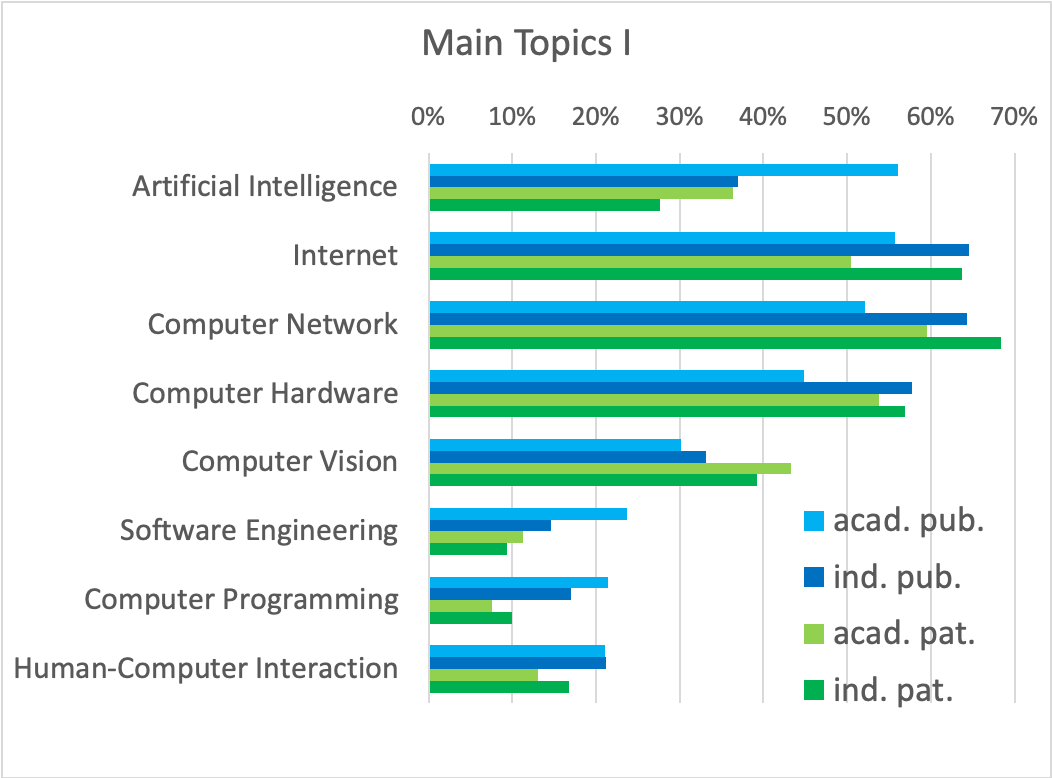}
\endminipage\hfill
\minipage{0.495\textwidth}
  \includegraphics[width=\linewidth]{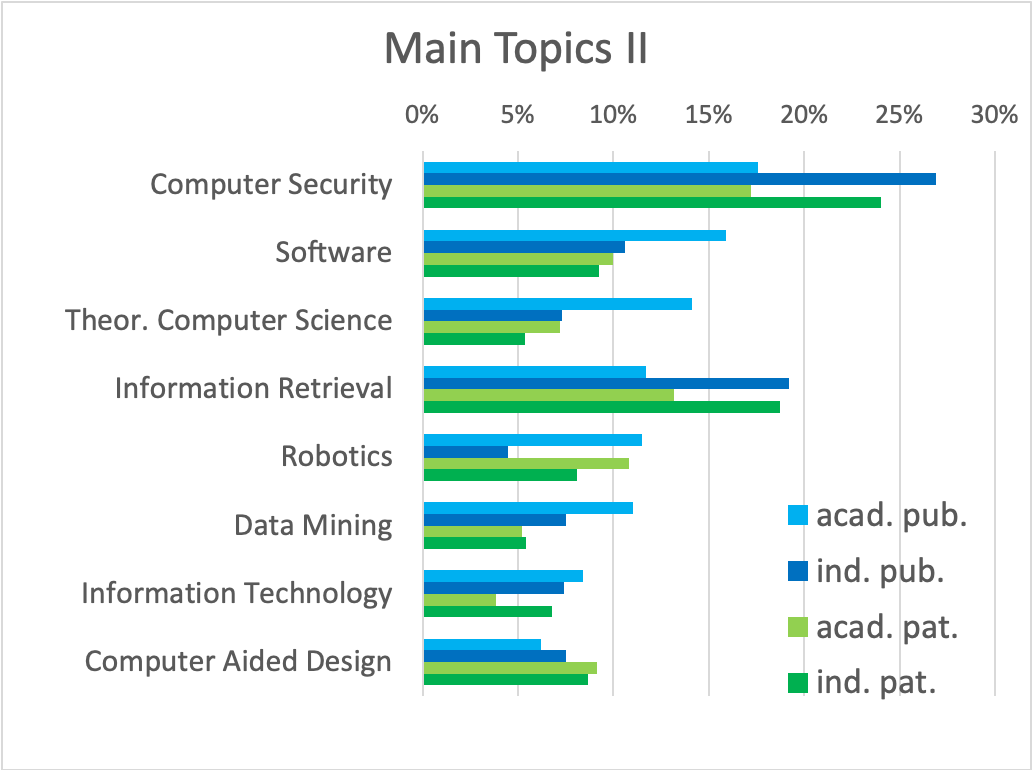}
\endminipage\hfill
 \caption{Distribution of the main topics in academia and industry.}\label{fig:mt}
\end{figure*}

AIDA was specifically designed for analysing the evolution of research topics across academia, industry, government facilities, and other institution. 
For instance, Figure~\ref{fig:mt} shows an example of the most frequent 16 high-level topics and reports the relevant percentage of academic publications, industry publications, academic patents, and industrial patents. Some topics, such as Artificial Intelligence and Theoretical Computer Science, are mostly addressed by academic publications. Others, e.g., Computer Security, Computer Hardware, and Information Retrieval, attract a stronger interest from the industry. The topics which are mostly associated with patents are Computer Networks, Internet, and  Computer Hardware. The overall sum of percentages for a given category on a certain topic may be more than 100\% because each document may be associated with multiple topics.

The current version only associates the affiliation type with about 5.1M out of 21M articles and 5.6M out of the 8M patents. 
The missing documents that could not be typed either have affiliations that are not present on GRID, or they were not correctly mapped to the relevant GRID ids in the original data, for instance, because it was not possible to identify the institution when parsing the article. 
This is an exemplary case of the incompleteness problem which ultimately affects all scholarly knowledge graphs: a field that could enable crucial analyses exists only for about 25\% of the documents.

This pragmatic scenario motivated us to investigate the best models for link prediction that could be applied on AIDA+MAG and by extensions on other knowledge graphs that suffer from similar issues.

AIDA, as many KGs in the scholarly domain, describes millions of documents according to a relatively small set of categories, making this task quite challenging for two main reasons. First, there is an abundance of N to M relations with N $\gg$ M, a situation that is not well handled by the current solutions, as we will further discuss in the next section. In addition, the number of items makes computationally unfeasible the adoption of embeddings with large dimensions. Therefore, it is critical to develop a solution that is able to both handle the N $\gg$ M issue and work well with limited dimensions. These considerations led to designing the novel Trans4E model presented in this paper.

Since we also wanted to test our solution on a field with different characteristics, we included in our analysis also the Fields of Study\footnote{Fields of Study - \url{https://academic.microsoft.com/topics/}}, which are terms from an in-house taxonomy used by MAG to index research papers. 
While we found that the topics from CSO typically produce a better representation of the domain of Computer Science~\cite{salatino2020computer} and thus are more apt for the analyses that focus on this discipline, FoS has the advantage of covering all research fields. In the context of AIDA this is particularly  useful for characterizing industrial sectors and companies that are interested in Computer Science applications in other fields such as Medicine, Chemistry, Geology, and Physics. Furthermore, FoS is a very interesting case study for a variety of reasons. First, its purpose is the description of the knowledge in the articles rather than their objective characteristics, such as the list of authors or the venues. Second, it is already present in the vanilla MAG, allowing us to experiment on a well established KG in this space. 
Finally, the quality of the topics for a specific document depends on the length and style of the abstract, which is typically parsed by an online PDF file, sometimes leading to  mistakes and missing content. 
Indeed, documents that are associated with short or incorrectly parsed abstracts are usually tagged with very few topics. 
Considering the relevant entities in the graph such as authors, venues, and references, may enable to identify the missing topics. 

Completing  the Fields of Study is critical for supporting deeper analyses of the trends of multiple disciplines. 
Besides, the same approach may be adopted to support the integration of documents from other knowledge graphs that lack abstracts (e.g., DBLP\footnote{DBLP - \url{https://dblp.org/}}, OpenCitations\footnote{OpenCitations - \url{https://opencitations.net/}}). For the sake of simplicity, in the rest of the paper we will refer to the Fields of Study from MAG in AIDA+MAG simply as \textit{topics}.

\subsection{Limitations of KGEs for Link Prediction}
To improve the number of documents in AIDA+MAG characterized according to their affiliation types, we tried four well-know embedding models: TransE, RotatE, ComplEx and QuatE. 
Their performance on this KG was not particularly good, in particularly when using small embedding vectors.

A systematic analysis showed that this was due to the characteristics of the \textit{hasGRIDType} relation: a N to M relation where N is large and M small, as in M $=$ 7: 'Education', 'Government', 'Company', 'Healthcare', 'Facility','Nonprofit', 'Other'.
Indeed, for any triple in the form  ($h,\textit{hasGRIDType},t$), there is a large number \emph{N} of $h$ entities (papers) given a specific tail $t$ (types), and a small number \emph{M} of $t$ (types) given a specific head $h$ (papers).
The same applies for the triples with the \textit{hasTopic} relation that associates articles with topics. 



The TransE model has some issues when handling these kinds of relations. 
Let us consider N entities $h$ (papers) typed as `Education' $(h, \textit{hasGRIDType}, \textit{Education})$. 
According to the TransE formulation
\ref{eq:tran}
(see Equation~\ref{eq:tran}), this case can be formalized as:
\begin{align} 
\begin{cases}
    \mathbf{h}_1 + \mathbf{r} = \mathbf{t},\\
    \mathbf{h}_2 + \mathbf{r} = \mathbf{t},\\
    \vdots \\
    \mathbf{h}_N + \mathbf{r} = \mathbf{t}.
\end{cases}
\label{trn2}
\end{align} 

Since the relation $\mathbf{r}$ is always $\textit{hasGRIDType}$ and the tail $\mathbf{t}$ always $\textit{Education}$, the formulation of the model enforces to have $\mathbf{h}_1 = \mathbf{h}_2 = \ldots =\mathbf{h}_N$ for N number of papers. 
This is an issue since the embedding vectors of all the entities appearing in the head will be very similar.
Consequently, the model may be unable to distinguish among them, resulting in poor performance. 
Therefore, the TransE model is more suitable for N to M relations in which N and M are both small.

RotatE suffers from the same issue.
According to the RotatE formulation, we have:
\begin{align} 
\begin{cases}
    \mathbf{h}_{1i} \circ \mathbf{r}_i = \mathbf{t}_i,\\
    \mathbf{h}_{2i} \circ \mathbf{r}_i = \mathbf{t}_i,\\
    \vdots\\
    \mathbf{h}_{Ni} \circ \mathbf{r}_i = \mathbf{t}_i,
    \qquad
     i = 1,\ldots,d.
\end{cases}
\label{rot}
\end{align} 

This shows that in each element of the embedding vector (indexed as $i$), there is only one option for embeddings for any head for a given tail. 
Therefore, in the case of the \textit{hasGRIDType} relation (also for \textit{hasTopic}), RotatE lacks the capacity to distinguish well among the research papers in the head ($h$). In conclusion, a larger vector space appears to be crucial to properly represent these kinds of relations and perform high-quality link prediction on AIDA and similar scholarly
knowledge graphs. 

Other models, such as ComplEx and QuatE, suffer from two major issues when applied on KG with high number of entities: a)  since they use 1 to $K$ negative sampling \cite{lacroix2018canonical} (where $K$ is the total number of entities in a KG), in the case of N to M relations where N $\gg$ M, a substantial portion of the samples are positive but they are used as negative samples; b) high computation costs also resulted from using $K$ negative sampling.






\section{Methodology}
\label{sec:methodology}

\subsection{The Trans4E Model}

Trans4E is a novel KGE model designed to effectively handle KGs which include N to M relations with N$\gg$M.  
In this section, we show that the capacity of this model for a given relation (e.g., \textit{hasGRIDType}, \textit{hasTopic}) and the corresponding tail entity (e.g., \textit{type} or \textit{topic}) is $8^d$, which allows to generate a distinct vector for each entity (e.g., a specific paper) even when using small embedding dimensions. 

Here we introduce the core formulation of the score function of Trans4E. 





Trans4E maps the entities of the graph via relations in Quaternion vector space $\mathbb{H}^d$. 
Concretely, given a triple of the form $(h,r,t)$, our model follows the following steps: 

\begin{itemize}
    \item[(a)] The head entity vector ($\mathbf{h} \in \mathbb{H}^d$) is rotated by $\mathbf{r}_\theta$ degrees in quaternion space i.e.~$\mathbf{h}_{\theta_r} = \mathbf{h} \otimes  \mathbf{r}_{\theta}.$ $\otimes$ is an element-wise Hamilton product between two quaternion vectors. 
    \item[(b)] The rotated head i.e.~$\mathbf{h}_{\theta_r}$ is translated by the relation embedding vector $\mathbf{r}$ to get $\mathbf{h}_r = \mathbf{h}_{\theta_r} + \mathbf{r}.$
    \item[(c)] The translated head embedding vector should meet
    the tail embedding vector i.e.~ $\mathbf{h}_r \approx \eta_h \otimes \mathbf{t}$ for a positive sample ($h,r,t$). $\mathbf{t} \in \mathbb{H}^d$. However, there is a possibility that the transformed vector of the head is not exactly
    meeting the tail. In order to solve this problem, we could use  $\eta_h = [\eta_{h1}, \ldots, \eta_{hd}] \in \mathbb{H}^d$, which is a mapping regularizer. 
\end{itemize}

%
%

Following the mentioned steps, we define the score function as:

\begin{equation}
\label{eq:score}
    f(\mathbf{h}, \mathbf{r}, \mathbf{t}) =- \|\mathbf{h}_r - \eta_h \otimes \mathbf{t}\|.
\end{equation}

The score function returns a low value if the triple is false i.e.~$\mathbf{h}_r \neq \eta_h \otimes \mathbf{t}$ and returns high value (close to zero) if the triple is true i.e.~$\mathbf{h}_r \approx \eta_h \otimes \mathbf{t}.$ In this way, we measure the plausibility of each triple $(h,r,t)$.

In addition, two regularized versions of the Trans4E model are also made available.
The first is Trans4EReg1, which is the regularized version of Trans4E where a relation-specific head rotation and the tail mapping regularizer are used. 
The second is Trans4EReg2, which is a regularized version of Trans4E with a relation-specific rotation on the tail side (in addition to the relation-specific head rotation and the tail mapping regularizer).

\subsection{Link Prediction on N to M Relations}
Here we show that Trans4E provides a higher capacity with fewer limitations than other models. 

Given a relation ${r}$ (e.g., \textit{hasGRIDType}) and a tail ${t}$ (e.g., 'Education'), the following constraints are applied for each of the resulting triples:

\begin{align}
\begin{cases}
    \mathbf{h}_{1\theta_{ri}} + \mathbf{r}_i = \eta_{h_{1i}}\otimes \mathbf{t}_i,\\
    \mathbf{h}_{2\theta_{ri}} + \mathbf{r}_i = \eta_{h_{2i}}\otimes \mathbf{t}_i,\\
    \vdots\\
    \mathbf{h}_{N\theta_{ri}} + \mathbf{r}_i = \eta_{h_{Ni}} \otimes \mathbf{t}_i,
    \qquad
     i = 1,\ldots,d.
\end{cases}
\label{trn}
\end{align}

We can rewrite the Hamilton product as 4-dimensional matrix-vector product:
\begin{equation}
\begin{split}
&\mathbf{h}_{\theta_{ri}} = \mathbf{h}_i \otimes  \mathbf{r}_{\theta_i} =\\
&\begin{bmatrix} a_{r_\theta}&-b_{r_\theta}&-c_{r_\theta}&-d_{r_\theta}\\
b_{r_\theta}&a_{r_\theta}&-d_{r_\theta}&c_{r_\theta}\\
c_{r_\theta}&d_{r_\theta}&a_{r_\theta}&-b_{r_\theta}\\
-d_{r_\theta}&-c_{r_\theta}&b_{r_\theta}&a_{r_\theta} \end{bmatrix} \begin{bmatrix} a_h\\b_h\\c_h\\d_h\end{bmatrix} = \mathcal{H}_i \vec{h}_i.
\end{split}
\end{equation}

Without loss of generality, we assume that the embedding of the relation translation $\mathbf{r}_i$ is zero and $\eta_{h_{pi}}$ is a real value. 
In this way, we can write the above system of equations in the following form:

\begin{align}
\begin{cases}
    \mathcal{H}_{i}\vec{h}_{1i} = \eta_{h_{1i}} \vec{t}_i \\
    \mathcal{H}_{i}\vec{h}_{2i} = \eta_{h_{2i}} \vec{t}_i \\
    \vdots\\
    \mathcal{H}_{i}\vec{h}_{1i} = \eta_{h_{Ni}} \vec{t}_i, 
    \qquad
    i = 1,\ldots,d.
\end{cases}
\end{align}

Note that the matrix $\mathcal{H}_{i}$ is $4\times4$ and has 4 distinct eigenvalues/eigenvectors. 
Therefore, we can write $\mathcal{H}_i\vec{h}_{pi} = \lambda_{h_{pi}} \vec{h}_{pi} =\eta_{h_{pi}} \vec{t}_{pi}.$ If $\lambda_{h_{pi}} = \eta_{h_{pi}}$, then the $i$th dimension of the head and tail vectors will be same, otherwise, they will be different. 
Therefore, we will have $4 \, (\text{number of distinct eigenvectors}) \times 2 \, = 8$
various options in each dimension to be assigned to the head entity vector. 
The multiplication by 2 is due to the two possible cases, one for the equality of the head and the tail and the other for their inequality. 


Because we use $d$ dimensional vectors, we have $8^d-1$ possible distinct vectors to be assigned to the entities appearing in the head (e.g., articles in AIDA). 
As a result, the capacity of the model becomes $8^d-1$, which provides a larger space than the TransE and RotatE models. 
In Section \ref{sec:evaluation}, we will show the advantages of this solution by comparing it against alternative models. 

\section{Evaluation}
\label{sec:evaluation}

We compared Trans4E against four alternative embedding models: TransE, RotatE, ComplEX, and QuatE. 

\subsection{Evaluation Datasets}

We ran the experiments on a portion of the knowledge graph \textit{AIDA+MAG} including 68,906 entities and 180K triples.  
Specifically, we considered the following entities:
publication IDs, authors, affiliation organizations, topics, publication types, conference editions, conference series, journals, years, countries, and references. 


%
%

In this subset, the \emph{hasGRIDType} relation includes about 5k entities (research papers) in the head position and 7 entities as tail (`Education', `Company', `Government', `Healthcare', `Nonprofit', `Facility', and `Other').

Regarding the \emph{hasTopic} relation, the highest number of research articles associated to a topic is 4,659, while the highest number of topics associated to research articles is only 13. 



We split the datasets into train (80\%), test (10\%), and validation (10\%) sets.
Additionally, we evaluated the performance of our model on four benchmarks: FB15K (14,951 entities and 1,345 relations), FB15k-237 (14,451 entities and 237 relations), WN18 (40,943 entities and 18 relations), and WN18RR (40,943 entities and 13 relations).





\subsection{Evaluation Criteria}
In this section we discuss the  criteria that we considered for the evaluation.

\textbf{Performance Metrics}.
The standard evaluating metrics for the performance of KGEs are:
Mean Rank (MR), Mean Reciprocal Rank (MRR) and Hits@k (k=1, 3, 10) \cite{wang2017knowledge}.  

MR is the average rank of correct triples in the test set. In order to compute it, we generate two sets of triples, $S_h = {(h,r,?)}$ and $S_t = {(?,r,t)}$, by corrupting each test triple ($h,r,t$).  After this step, the scores of all the triples in $S_h, S_t$ are computed and the triples are sorted. The rank ($r_h, r_t$) of the original triple (i.e.\ $(h,r,t)$) is then computed in both sets $S_h$, and $S_t$. For any triple, $r_h$ is the notation for the right ranks and $r_t$  for the left ranks. 
The rank of the example triple of ($h,r,t$) is computed as $rank = \frac{r_h + r_t}{2}$. 
If we assume $rank_i$ to be the rank of the $i-$th triple in the test set obtained by a KGE model, then the MR and the MRR are obtained as follows:
\begin{equation*}
    MR = \sum_i rank_i,
\end{equation*}

\begin{equation*}
    MRR = \sum_i \frac{1}{rank_i}.
\end{equation*}

For the evaluation on \emph{hasGRIDType} and \emph{hasTopic} relations, we only corrupted the tail of the relations and replaced it with all the entities in the KG.

The $Hits@K$, for k = 1, 3, 10 \dots, is one of the standard link prediction measurements. 
By considering the percentage of the triples for which $rank_i$ is equal or smaller than \textit{k}, we computed the $Hits@K$. 
MR, the average MRR, Hits@1, Hits@3, and Hits@10 are reported in Tables 2-6. 

\begin{table*}[h!]
\caption{Performance of KGEs on AIDA for Dimension 5}
\label{table:aida2}
\begin{adjustbox}{width=\textwidth,center}
\begin{tabular}{llllllllllll}
\hline
{ Model Type}  & { }                             & \multicolumn{5}{l}{{ hasTopic}}                                                                                                                                                                                                & \multicolumn{5}{l}{{ hasGRIDType}}                                                                                                                                                                                                                       \\ \hline
{ }            & { MR}                           & { MRR}                           & { Hits@1}                        & { Hits@3}                        & { Hits@10}                       & { } & { MR}                        & { MRR}                           & { Hits@1}                        & { Hits@3}                        & { Hits@10}                       \\
{ TransE}      & { 3785}                         & { 0.031}                         & { 0.006}                         & { 0.027}                         & { 0.071}                         & { } & { 6}                         & { 0.658}                         & { 0.500}                         & { 0.771}                         & { 0.970}                         \\ \hline
{ RotatE}      & { 4749}                         & { 0.036}                         & { 0.000}                         & { 0.001}                         & { 0.008}                         & { } & { 38}                        & { 0.472}                         & { 0.000}                         & { 0.000}                         & { 0.001}                         \\ \hline
{ QuatE}       & { 4862}                         & { 0.066}                         & { 0.021}                         & { 0.066}                         & { 0.151}                         & { } & { 159}                       & { 0.252}                         & { 0.166}                         & { 0.271}                         & { 0.431}                         \\ \hline
{ ComplEx}     & { 3726}                         & { 0.044}                         & { 0.003}                         & { 0.042}                         & { 0.111}                         & { } & { 6}                         & { 0.429}                         & { 0.001}                         & { 0.838}                         & { 0.931}                         \\ \hline \hline
{ Trans4EReg1}  & { 3007}                         & \cellcolor[HTML]{EBEBFF}{ 0.403} & \cellcolor[HTML]{EBEBFF}{ 0.325} & \cellcolor[HTML]{EBEBFF}{ 0.450} & \cellcolor[HTML]{EBEBFF}{ 0.531} & { } & \cellcolor[HTML]{EBEBFF}{ 1} & { 0.941}                         & { 0.915}                         & { 0.978}                         & { 0.995}                         \\ \hline
{ Trans4EReg2} & \cellcolor[HTML]{EBEBFF}{ 2047} & { 0.401}                         & \cellcolor[HTML]{EBEBFF}{ 0.325} & { 0.445}                         & { 0.528}                         & { } & \cellcolor[HTML]{EBEBFF}{ 1} & \cellcolor[HTML]{EBEBFF}{ 0.956} & \cellcolor[HTML]{EBEBFF}{ 0.928} & \cellcolor[HTML]{EBEBFF}{ 0.985} & { 0.988}                         \\ \hline
{ Trans4E}     & { 2908}                         & { 0.089}                         & { 0.030}                         & { 0.083}                         & { 0.211}                         & { } & \cellcolor[HTML]{EBEBFF}{ 1} & { 0.900}                         & { 0.834}                         & { 0.965}                         & \cellcolor[HTML]{EBEBFF}{ 0.998} \\ \hline

\end{tabular}
\end{adjustbox}
\end{table*}


\begin{table*}[h!]
\caption{Performance of KGEs on AIDA for Dimension 50}
\label{table:aida3}
\begin{adjustbox}{width=\textwidth,center}
\begin{tabular}{llllllllllll}
\hline
{ Model Type}  & { }                             & \multicolumn{5}{l}{{ hasTopic}}                                                                                                                                                                                                & \multicolumn{5}{l}{{ hasGRIDType}}                                                                                                                                                                                                                       \\ \hline
{ }            & { MR}                           & { MRR}                           & { Hits@1}                        & { Hits@3}                        & { Hits@10}                       & { } & { MR}                        & { MRR}                           & { Hits@1}                        & { Hits@3}                        & { Hits@10}                       \\
{ TransE}      & { 3903}                         & { 0.135}                         & { 0.043}                         & { 0.126}                         & { 0.355}                         & { } & \cellcolor[HTML]{EBEBFF}{ 1} & { 0.859}                         & { 0.769}                         & { 0.944}                         & \cellcolor[HTML]{EBEBFF}{ 1.000} \\ \hline
{ RotatE}      & { 3890}                         & { 0.155}                         & { 0.057}                         & { 0.144}                         & { 0.411}                         & { } & \cellcolor[HTML]{EBEBFF}{ 1} & { 0.891}                         & { 0.823}                         & { 0.970}                         & \cellcolor[HTML]{EBEBFF}{ 1.000} \\ \hline
{ QuatE}       & \cellcolor{blue!8}{ 1693}                         & { 0.093}                         & { 0.057}                         & { 0.106}                         & { 0.165}                         & { } & { 1718}                      & { 0.096}                         & { 0.062}                         & { 0.116}                         & { 0.148}                         \\ \hline
{ ComplEx}     & { 7279}                         & { 0.081}                         & { 0.036}                         & { 0.093}                         & { 0.167}                         & { } & { 700}                       & { 0.896}                         & { 0.869}                         & { 0.919}                         & { 0.939}                         \\ \hline \hline
{ Trans4EReg1}  & { 2424} & { 0.379}                         & { 0.300}                         & { 0.416}                         & \cellcolor[HTML]{EBEBFF}{ 0.515} & { } & { 117}                       & { 0.907}                         & { 0.856}                         & { 0.947}                         & { 0.991}                         \\ \hline 
{ Trans4EReg2} & { 3250}                         & \cellcolor[HTML]{EBEBFF}{ 0.394} & \cellcolor[HTML]{EBEBFF}{ 0.327} & \cellcolor[HTML]{EBEBFF}{ 0.429} & { 0.507}                         & { } & \cellcolor[HTML]{EBEBFF}{ 1} & \cellcolor[HTML]{EBEBFF}{ 0.959} & \cellcolor[HTML]{EBEBFF}{ 0.928} & \cellcolor[HTML]{EBEBFF}{ 0.990} & \cellcolor[HTML]{EBEBFF}{ 1.000} \\ \hline
{ Trans4E}     & { 3842}                         & { 0.158}                         & { 0.053}                         & { 0.154}                         & { 0.416}                         & { } & \cellcolor[HTML]{EBEBFF}{ 1} & { 0.866}                         & { 0.790}                         & { 0.931}                         & \cellcolor[HTML]{EBEBFF}{ 1.000} \\ \hline
\end{tabular}
\end{adjustbox}
\end{table*}


\begin{table*}[h!]
\caption{Performance of KGEs on AIDA for dimension 500.}
\label{table:aida4}
\begin{adjustbox}{width=\textwidth,center}
\begin{tabular}{llllllllllll}
\hline
{ Model Type}  & { }                             & \multicolumn{5}{l}{{ hasTopic}}                                                                                                                                                                                                & \multicolumn{5}{l}{{ hasGRIDType}}                                                                                                                                                                                                                       \\ \hline
{ }            & { MR}                           & { MRR}                           & { Hits@1}                        & { Hits@3}                        & { Hits@10}                       & { } & { MR}                        & { MRR}                           & { Hits@1}                        & { Hits@3}                        & { Hits@10}                       \\
{ TransE}      & { 3982}                         & { 0.400}                         & { 0.294}                         & { 0.462}                         & { 0.592}                         & { } & \cellcolor[HTML]{EBEBFF}{ 1} & \cellcolor[HTML]{EBEBFF}{ 0.968} & \cellcolor[HTML]{EBEBFF}{ 0.944} & { 0.990}                         & \cellcolor[HTML]{EBEBFF}{ 1.000} \\ \hline
{ RotatE}      & { 4407}                         & \cellcolor[HTML]{EBEBFF}{ 0.433} & { 0.332}                         & \cellcolor[HTML]{EBEBFF}{ 0.492} & { 0.622}                         & { } & \cellcolor[HTML]{EBEBFF}{ 1} & { 0.953}                         & { 0.933}                         & { 0.975}                         & { 0.996}                         \\ \hline
{ QuatE}       & \cellcolor[HTML]{EBEBFF}{ 1353} & { 0.426}                         & \cellcolor[HTML]{EBEBFF}{ 0.341} & { 0.472}                         & { 0.581}                         & { } & \cellcolor[HTML]{EBEBFF}{ 1} & { 0.957}                         & { 0.928}                         & { 0.983}                         & { 0.998}                         \\ \hline
{ ComplEx}     & { 5855}                         & { 0.099}                         & { 0.077}                         & { 0.109}                         & { 0.129}                         & { } & { 1566}                      & { 0.566}                         & { 0.531}                         & { 0.596}                         & { 0.609}                         \\ \hline \hline
{ Trans4EReg1}  & { 2040}                         & { 0.402}                         & { 0.295}                         & { 0.466}                         & { 0.604}                         & { } & { 233}                       & { 0.910}                         & { 0.882}                         & { 0.937}                         & { 0.944}                         \\ \hline
{ Trans4EReg2} & { 1942}                         & { 0.424}                         & { 0.325}                         & { 0.482}                         & { 0.602}                         & { } & { 34}                        & { 0.955}                         & { 0.931}                         & { 0.978}                         & { 0.990}                         \\ \hline
{ Trans4E}     & { 3904}                         & { 0.426}                         & { 0.318}                         & \cellcolor[HTML]{EBEBFF}{ 0.492} & \cellcolor[HTML]{EBEBFF}{ 0.628} & { } & \cellcolor[HTML]{EBEBFF}{ 1} & \cellcolor[HTML]{EBEBFF}{ 0.968} & \cellcolor[HTML]{EBEBFF}{ 0.944} & \cellcolor[HTML]{EBEBFF}{ 0.995} & { 0.998}                         \\ \hline

\end{tabular}
\end{adjustbox}
\end{table*}

\begin{figure*}[h!]
\centering
\minipage{0.495\textwidth}
  \includegraphics[width=\linewidth]{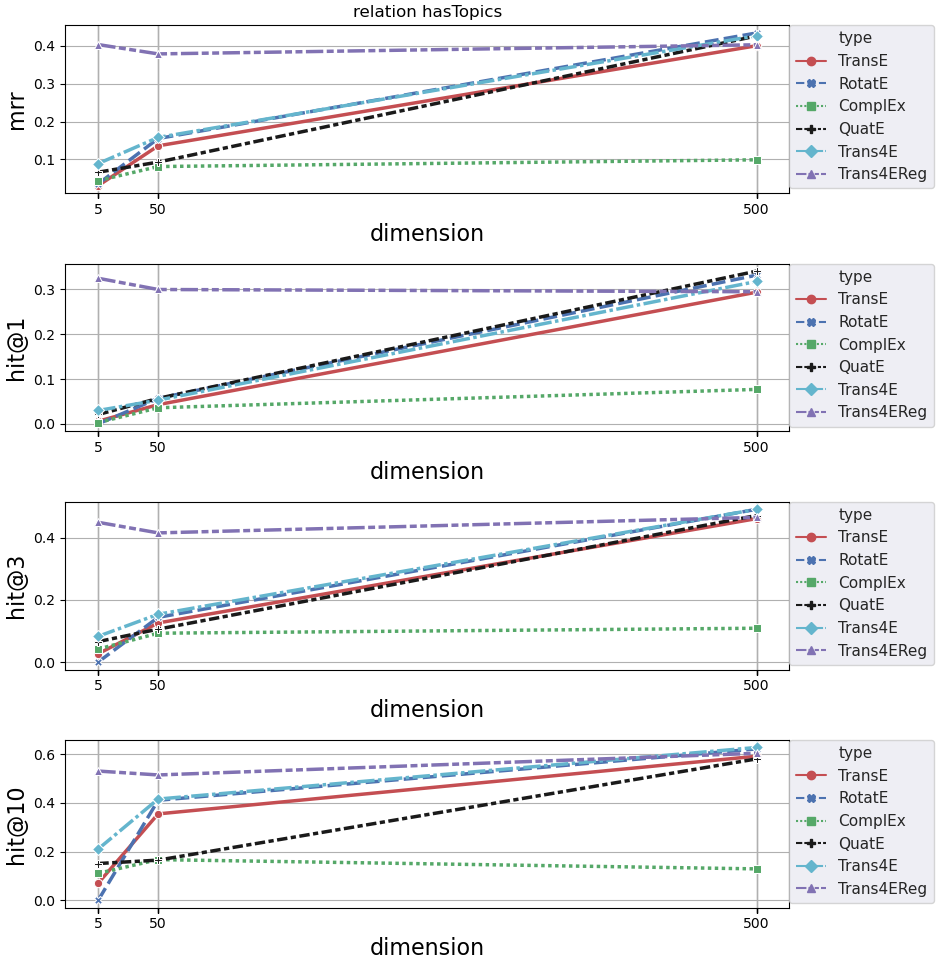}
 \caption{hasTopic for dimension 5,50 and 500}\label{fig:hasTopics}
\endminipage\hfill
\minipage{0.495\textwidth}
  \includegraphics[width=\linewidth]{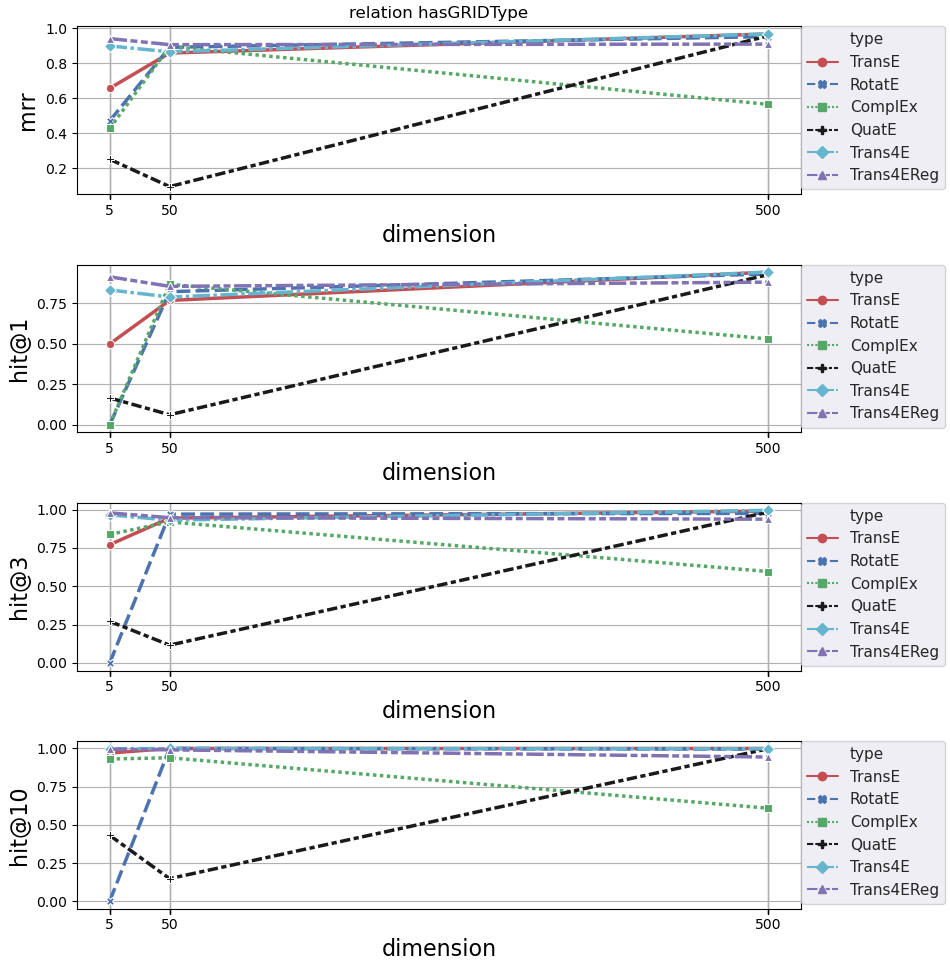}
  \caption{hasGRIDType for dimension 5,50 and 500}\label{fig:hasTypes}
\endminipage\hfill
\end{figure*}

\textbf{Dimension and KG Scale.}
Although the performance measures of a machine learning model are important criteria for evaluation, the dimension of the embedding vectors is specifically important for KGE models, which are supposed to be used in the real-world large-scale KGs. Indeed, an embedding with very large dimensions may be unfeasible in most practical settings. 

Therefore, we compared the performances of our model against state-of-the art models in a very low dimensional embedding.
This was done to simulate a real-world application of KGEs on large scale KGs. Indeed, models which obtain satisfactory performances on a portion of a graph using a small vector size should also perform well when adopting a higher dimension on a larger portion of the same graph \cite{nayyeri2020fantastic,goyal2018graph}. 

\subsection{Hyperparameter Setting}
The development environment of our model is PyTorch\footnote{PyTorch - \url{https://pytorch.org/}}. 
In the experiments, we reshuffled the training set in each epoch, and generated 16 mini batches on the reshuffled samples. 
To determine the performances of our model in high and low dimensions, the embedding dimension ($d$) was set to $\{5,50,500\}$ in the experiments.  
The batch size ($b$) is considered as $\{256,512\}$, the fixed margin $\gamma$ is  $\{2,3,4,5,10,15,20,30\}$ and learning rate as $\{0.001, 0.01,0.05,0.1\}$ with a negative sample of 10.
$L_{2}$ regularization coefficient is $\{0.000005,0.0000005\}$ for the models QuatE, Trans4EReg1, and Trans4EReg2. 
The best hyperparameter combination for Trans4E and Trans4EReg2 is $b$ = 256, $lr=0.1, \gamma =20$ and for Trans4EReg1  is $b$ = 256, $lr=0.001, \gamma =20$, and $d$ = 500 for all the models. For the regularized versions $\lambda$ =0.000005.

\subsection{Results and Discussions}
In this section, we present the results of our experiments. 
Specifically, Section \ref{sec:kgcaida} reports the results of the evaluation regarding  the graph completion on AIDA+MAG. Section \ref{sec:lpbd} compares the performance of Trans4E and several alternatives on a set of four standard benchmarks (FB15k, FB15k-237, WN18, andWN18RR). Section 5.4.3 investigates the representation of the research topics and shows a study of the distribution of their embedding vectors.


\subsubsection{Knowledge Graph Completion in AIDA+MAG}
\label{sec:kgcaida}
In this section we evaluate the performance of Trans4E versus alternative methods in completing the two relations \emph{hasGRIDType} and \emph{hasTopic} in AIDA+MAG.

Specifically, we compared Trans4E with TransE, RotatE, QuatE and ComplEx. We also included Trans4EReg1 and Trans4EReg2, the two regularized versions previously defined in Section 4.1. 


Table \ref{table:aida2} reports the performances of the seven models for dimension 5.
Trans4EReg1 clearly outperforms all the other models for the \emph{hasTopic} relations.
Trans4EReg2 obtains the second-best performance.
For instance, when considering the \emph{hasTopic} relation, Trans4EReg1 and Trans4EReg2 yield 32.5\% in Hits@1 while all the other solutions obtain less than 3\%. 
For the \emph{hasGRIDType} relations Trans4EReg2 outperforms all the others with a 92.8\% in Hits@1. Moreover, Trans4EReg1, yields 91.5\% in Hits@1 and Trans4E 83.4\%, while the best of the other models is TransE with 50.0\%.




RotatE performed surprisingly poorly on both the \emph{hasTopic} and \emph{hasGRIDType} relations, yielding 0\% in Hits@1.
It should be noted that during testing, for a test triple ($p, hasGRIDType, t$), we replaced the tail $t$ with all the entities in the graph and ranked the actual ones against the corrupted triples. As a result, RotatE (in dimension 5) does not rank any type entity even among the top 10 occurrences. This means that non-type entities in the corruption process are ranked higher than the typed entities. This is related to the limited solution space of the RotatE model, which is also discussed in~\cite{iswcfantastic}.




The overall accuracy for \emph{hasGRIDType} is typically higher than \emph{hasTopic}.
For instance, Trans4EReg1 yields a Hits@10 of 99.5\% for  \emph{hasGRIDType} and  53.1\% for \emph{hasTopic}.
This is mainly due to the fact that the number of entities to be considered for \emph{hasTopic} is much higher than that for \emph{hasGRIDType}.

Overall, Trans4EReg1 seems to be the most suitable model for addressing large-scale KGs, where increasing the dimension of the model is too costly in computational terms.

\begin{table*}[t]
\caption{Performance of KGEs on FB15K and WN18.}
\label{table:expr4}
\begin{adjustbox}{width=\textwidth,center}
\begin{tabular}{llllllllllll}
\hline
Model Type                       &  & \multicolumn{5}{l}{FB15k}         & \multicolumn{5}{l}{WN18}      \\ \cline{1-1} \cline{3-12} 
  &      MR & MRR & Hits@1 & Hits@3 & Hits@10 &  &MR & MRR & Hits@1 & Hits@3 & Hits@10 \\ 
TransE            &  -- &  0.463&  0.297& 0.578&  0.749& & -- & 0.495& 0.113& 0.888& 0.943 \\ \hline
RotatE            &   40&  \cellcolor{blue!8}0.797&   \cellcolor{blue!8}0.746 &  0.830& 0.884& & 309 & 0.949& 0.944& 0.952&0.959 \\ \hline
QuatE           &   \cellcolor{blue!8}35&  0.742&  0.658& 0.805&  0.881& & 349 & 0.942& 0.927& 0.952& 0.960\\ \hline
\hline
Trans4E      &   47&  0.767&  0.681&  \cellcolor{blue!8}0.834&  \cellcolor{blue!8}0.892& & \cellcolor{blue!8}175 & \cellcolor{blue!8}0.950& \cellcolor{blue!8}0.944& \cellcolor{blue!8}0.953& \cellcolor{blue!8}0.960\\ \hline
\end{tabular}
\end{adjustbox}
\end{table*}

\begin{table*}[t]
\caption{Performance of KGEs on FB15K-237 and WN18RR.}
\label{table:expr5}
\begin{adjustbox}{width=\textwidth,center}
\begin{tabular}{llllllllllll}
\hline
Model Type                       &  & \multicolumn{5}{l}{FB15k-237}         & \multicolumn{5}{l}{WN18RR}      \\ \cline{1-1} \cline{3-12} 
  &      MR & MRR & Hits@1 & Hits@3 & Hits@10 &  &MR & MRR & Hits@1 & Hits@3 & Hits@10 \\ 
TransE            & 357 &  0.294&  --& --&  0.465& & 3384& 0.226& --& --& 0.501 \\ 
\hline
RotatE            &177&  \cellcolor{blue!8}0.338&  \cellcolor{blue!8}0.241&  \cellcolor{blue!8}0.375& \cellcolor{blue!8}0.533& & 3340 & \cellcolor{blue!8}0.476& \cellcolor{blue!8}0.428& \cellcolor{blue!8}0.492&0.571 \\ \hline
QuatE           &   170&  0.282&  0.178& 0.315&  0.501& & 2272 & 0.303& 0.179& 0.386& 0.530\\ \hline
\hline
Trans4E      &   \cellcolor{blue!8}158&  0.332&  0.236&  0.366&  0.527& & \cellcolor{blue!8}1755 & 0.469& 0.416& 0.487& \cellcolor{blue!8}0.577\\ \hline
\end{tabular}
\end{adjustbox}
\end{table*}


Table \ref{table:aida3} reports the performances of the models using dimensions 50.
Trans4EReg1 and Trans4EReg2 outperforms all the models with regards to the  \emph{hasTopic} by a considerable margin (up to 10\% improvement on Hits@10). When considering \emph{hasGRIDType}, Trans4EReg2 obtains the best performances in all metrics, folowed by Trans4EReg1 and RotatE. 
Due to the overfitting, the performance of Trans4EReg1 and Trans4EReg2 decreases as the dimension increases from 5 to 50. In fact, Trans4EReg1 and Trans4EReg2 with dimension 5 still outperforms all the models with dimension 50 in most of the metrics.



Table \ref{table:aida4} reports the experiments with a dimension of 500. For \emph{hasGRIDType}, Trans4E and TransE are comparable and obtain the best performances.
When considering \emph{hasTopic}, QuatE, RotatE, and Trans4E perform similarly well. Specifically, QuatE yields the best performance in Hits@1 (34.1\%), while Trans4E and RotatE perform best in Hits@3 (49.2\%), and Trans4E obtains the highest Hits@10 (62.8\%). 


\autoref{fig:hasTopics} and \ref{fig:hasTypes} summarize the performances of all the models for dimension 5, 50, and 500. Trans4EReg1 significantly outperforms all the models when using low dimensions and performs well also in high dimensions.   




\begin{figure}[h!]
\centering
  \includegraphics[width=\linewidth]{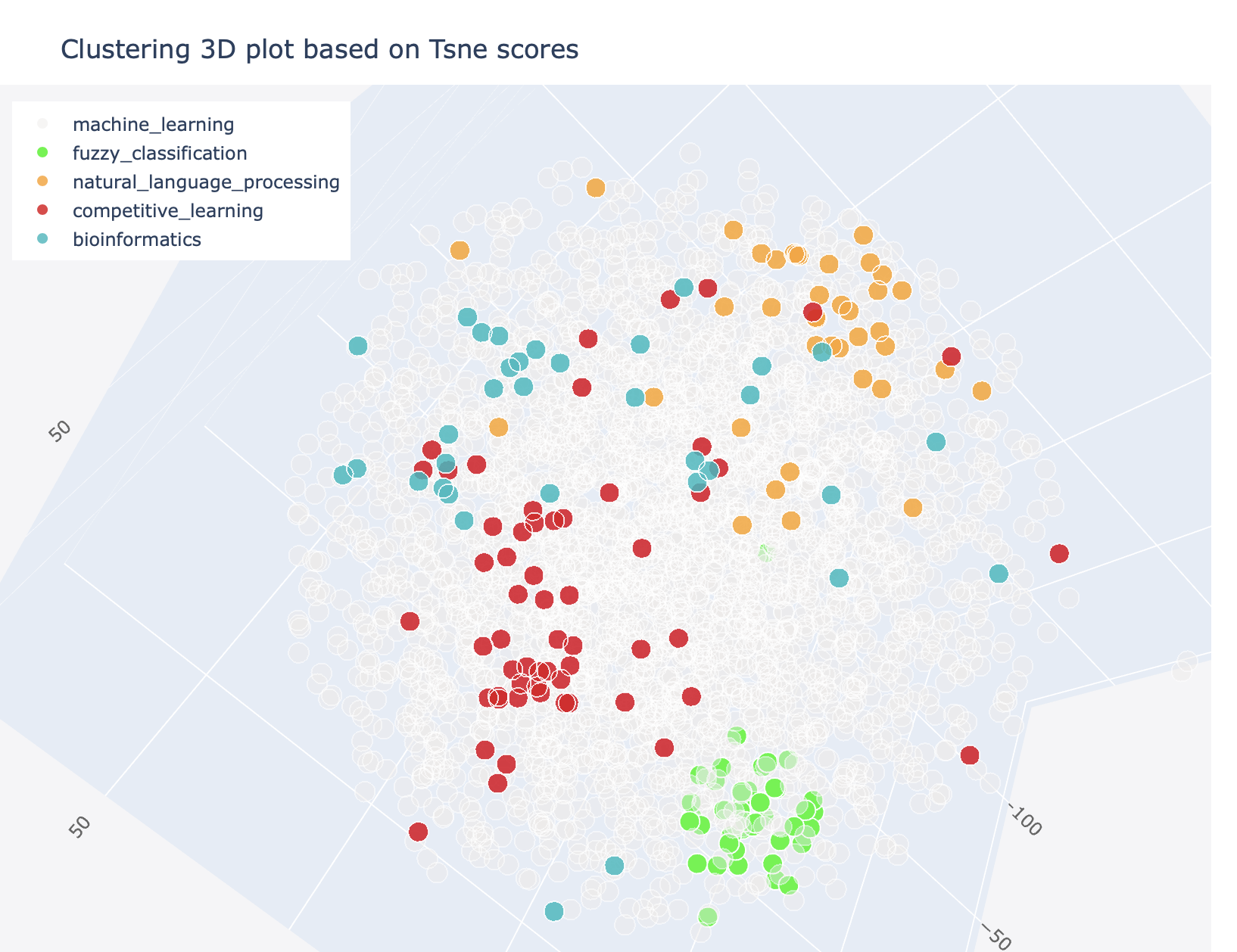}
 \caption{Distribution of the main topics in academia and industry.}
 \label{fig:cluster6b}
\end{figure}

\begin{figure*}[ht!]
\centering
    \includegraphics[trim={0cm 0 0cm 0},clip,width=.8\textwidth]{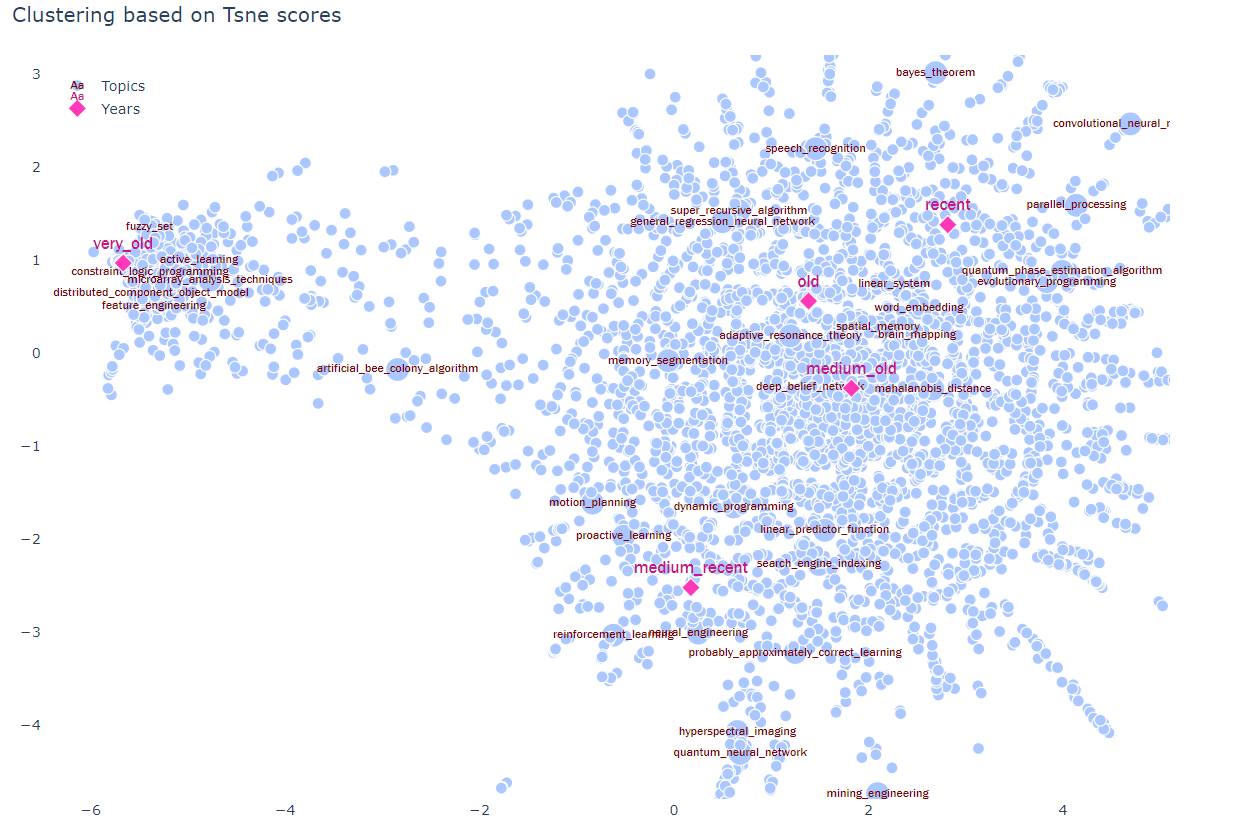}
    \caption{\textbf{Distribution of Topics w.r.t Years}. year >= 2015 is considered recent, year >= 2010 and year < 2015 are denoted as medium\_recent, year >= 2005 and year < 2010 are medium\_old, year>= 2000 and year < 2005 mean old, and
anything before 2000 is very\_old.}
    \label{fig:topicyear}
\end{figure*}

\subsubsection{Link Prediction on Benchmark Datasets}
\label{sec:lpbd}
We evaluated the performances of the Trans4E model against the competitors on a set of standard benchmark datasets with diverse relations 

(N to M relations where N and M are large, N and M are small, N$\gg$M and N$\ll$M).

Table \ref{table:expr4} and Table \ref{table:expr5} show the performances of the KGE models on the benchmark datasets FB15k, FB15k-237, WN18, and WN18RR.
Trans4E outperforms the other models in Hits@3 and Hits@10 in FB15k and WN18. 
It also obtains a significantly better MR on FB15k-237 and WN18RR. In FB15k, the Trans4E model outperforms all the other models when considering the Hits@3 and Hits@10.
In WN18, Trans4E outperforms TransE and QuatE, and obtains competitive results with respect to RotatE. 
To note that, these results are computed by running the models on the benchmark datasets using the best obtained hyperparameter settings where the dimension is 200, and with 20 negative samples using adversarial negative sampling \cite{sun2018rotate}. 
The results are comparatively close in the case of FB15k-237 and WN18RR, where Trans4E has a better performance in MR. 

Overall, the results show that our model outperforms other KGE models on N to M relations with N$\gg$M and provides competitive performance on KGs with diverse relations.

\subsubsection{Efficiency of the Embeddings}
To further investigate the representation of research topics with Trans4E, we analysed how the embeddings discriminate articles tagged with different topics.

Figure \ref{fig:cluster6b} shows the embeddings associated to the articles in AIDA+MAG in two dimensions.  In order to produce it, we first selected five major topics of the machine learning venues: ``fuzzy\_classification'', ``natural\_language\_processing'', ``competitive\_learning'', ``machine learning'',  and ``bioinformatics''. Then, we retrieved the embedding vectors of the papers tagged with those topics and visualized them by using T-SNE \cite{maaten2008visualizing}. 

We can appreciate how papers with the same topics tend to cluster together. 
For example, papers belonging to the ``fuzzy\_classification'' topic (green) lie within the same cluster. 
Note that papers in some topics such as ``bioinformatics'' may be associated to other topics as well (e.g.~a paper may be in ``bioinformatics'' and use ``fuzzy\_classification'' methods). This is why papers related to more general topics are distributed with a larger variance.


We further evaluated the ability of our model to properly distribute topics in the vector space based on their publication dates. 
In Figure \ref{fig:topicyear}, we illustrate the distribution of the learned vectors for the topics w.r.t their publishing years.
This shows that topics such as ``convolutional\_neural\_networks'', ``parallel\_processing'', and ``speech\_recognition'' are correctly identified to be hot topics for the corresponding years.

The topic ``word\_embedding'' lies in the border of recent and medium\_old period indicating that even if old is still lasting.
There is also a cluster of topics around the very\_old time period for which the corresponding vectors are very different from the ones in other time periods. A manual analysis revealed that most of them were mostly active before the year 2000.

\section{Conclusion and Future Work}
\label{sec:conclusion}
In this paper we presented Trans4E, a KGE model designed to provide link prediction for KGs that include N to M relations with N$\gg$M. 

Trans4E and its regularized versions (Trans4EReg1 and Trans4EReg2) have been applied on a real world case involving the Academic/Industry DynAmics Knowledge Graphs (AIDA) and the Microsoft Academic Graph (MAG).
The evaluation showed that Trans4E outperforms other approaches in the case of N to M relations with N $\gg$ M and obtains competitive results in all the other settings, in particular when using low embedding dimensions.
Hence, our approach appears to be an effective and generalizable solution able to achieve and sometimes improve state-of-the-art performance on many established benchmarks. In addition, it seems to be the most effective model when dealing with shallow classification schemes and using low embedding dimensions.





In future work we aim to perform link prediction on other relations in AIDA and MAG, with the aim  to release a new version of these KGs for the research community.
We also intend to use Trans4E for supporting a variety of other tasks involving scholarly KGs, such as trend detection, expert search, and  recommendation of articles and patents. Finally, we plan to investigate the application of Trans4E on KGs describing scientific knowledge, such as AI-KG~\cite{dessi2020ai} and ORKG~\cite{jaradeh2019open}.

\section*{Acknowledgements}
We gratefully acknowledge the support of NVIDIA Corporation with the donation of the Titan X GPU used for this research.
This work is supported by the EC Horizon 2020 grant LAMBDA (GA no.~809965), and the CLEOPATRA project (GA no.~812997).
\bibliographystyle{cas-model2-names}

\bibliography{bibliography}

\begin{thebibliography}{64}
\expandafter\ifx\csname natexlab\endcsname\relax\def\natexlab#1{#1}\fi
\providecommand{\url}[1]{\texttt{#1}}
\providecommand{\href}[2]{#2}
\providecommand{\path}[1]{#1}
\providecommand{\DOIprefix}{doi:}
\providecommand{\ArXivprefix}{arXiv:}
\providecommand{\URLprefix}{URL: }
\providecommand{\Pubmedprefix}{pmid:}
\providecommand{\doi}[1]{\href{http://dx.doi.org/#1}{\path{#1}}}
\providecommand{\Pubmed}[1]{\href{pmid:#1}{\path{#1}}}
\providecommand{\bibinfo}[2]{#2}
\ifx\xfnm\relax \def\xfnm[#1]{\unskip,\space#1}\fi
\bibitem[{Ammar et~al.(2018)Ammar, Groeneveld, Bhagavatula, Beltagy, Crawford,
  Downey, Dunkelberger, Elgohary, Feldman, Ha et~al.}]{ammar2018construction}
\bibinfo{author}{Ammar, W.}, \bibinfo{author}{Groeneveld, D.},
  \bibinfo{author}{Bhagavatula, C.}, \bibinfo{author}{Beltagy, I.},
  \bibinfo{author}{Crawford, M.}, \bibinfo{author}{Downey, D.},
  \bibinfo{author}{Dunkelberger, J.}, \bibinfo{author}{Elgohary, A.},
  \bibinfo{author}{Feldman, S.}, \bibinfo{author}{Ha, V.}, et~al.,
  \bibinfo{year}{2018}.
\newblock \bibinfo{title}{Construction of the literature graph in semantic
  scholar}.
\newblock \bibinfo{journal}{arXiv preprint arXiv:1805.02262} .
\bibitem[{Angioni et~al.(2020)Angioni, Salatino, Osborne, Recupero and
  Motta}]{angioni2020integrating}
\bibinfo{author}{Angioni, S.}, \bibinfo{author}{Salatino, A.A.},
  \bibinfo{author}{Osborne, F.}, \bibinfo{author}{Recupero, D.R.},
  \bibinfo{author}{Motta, E.}, \bibinfo{year}{2020}.
\newblock \bibinfo{title}{Integrating knowledge graphs for analysing academia
  and industry dynamics}, in: \bibinfo{booktitle}{ADBIS, TPDL and EDA 2020
  Common Workshops and Doctoral Consortium}, \bibinfo{organization}{Springer}.
  pp. \bibinfo{pages}{219--225}.
\bibitem[{Arora(2020)}]{arora2020survey}
\bibinfo{author}{Arora, S.}, \bibinfo{year}{2020}.
\newblock \bibinfo{title}{A survey on graph neural networks for knowledge graph
  completion}.
\newblock \bibinfo{journal}{arXiv preprint arXiv:2007.12374} .
\bibitem[{Belleau et~al.(2008)Belleau, Nolin, Tourigny, Rigault and
  Morissette}]{belleau2008bio2rdf}
\bibinfo{author}{Belleau, F.}, \bibinfo{author}{Nolin, M.A.},
  \bibinfo{author}{Tourigny, N.}, \bibinfo{author}{Rigault, P.},
  \bibinfo{author}{Morissette, J.}, \bibinfo{year}{2008}.
\newblock \bibinfo{title}{Bio2rdf: towards a mashup to build bioinformatics
  knowledge systems}.
\newblock \bibinfo{journal}{Journal of biomedical informatics}
  \bibinfo{volume}{41}, \bibinfo{pages}{706--716}.
\bibitem[{Bendre et~al.(2020)Bendre, Mar{\'\i}n and
  Najafirad}]{bendre2020learning}
\bibinfo{author}{Bendre, N.}, \bibinfo{author}{Mar{\'\i}n, H.T.},
  \bibinfo{author}{Najafirad, P.}, \bibinfo{year}{2020}.
\newblock \bibinfo{title}{Learning from few samples: A survey}.
\newblock \bibinfo{journal}{arXiv preprint arXiv:2007.15484} .
\bibitem[{Bonatti et~al.(2019)Bonatti, Decker, Polleres and
  Presutti}]{bonatti2019knowledge}
\bibinfo{author}{Bonatti, P.A.}, \bibinfo{author}{Decker, S.},
  \bibinfo{author}{Polleres, A.}, \bibinfo{author}{Presutti, V.},
  \bibinfo{year}{2019}.
\newblock \bibinfo{title}{Knowledge graphs: New directions for knowledge
  representation on the semantic web (dagstuhl seminar 18371)}, in:
  \bibinfo{booktitle}{Dagstuhl Reports}, \bibinfo{organization}{Schloss
  Dagstuhl-Leibniz-Zentrum fuer Informatik}.
\bibitem[{Bordes et~al.(2013)Bordes, Usunier, Garcia-Duran, Weston and
  Yakhnenko}]{bordes2013translating}
\bibinfo{author}{Bordes, A.}, \bibinfo{author}{Usunier, N.},
  \bibinfo{author}{Garcia-Duran, A.}, \bibinfo{author}{Weston, J.},
  \bibinfo{author}{Yakhnenko, O.}, \bibinfo{year}{2013}.
\newblock \bibinfo{title}{Translating embeddings for modeling multi-relational
  data}, in: \bibinfo{booktitle}{Advances in NIPS}.
\bibitem[{Bose et~al.(2019)Bose, Jain, Molino and Hamilton}]{bose2019meta}
\bibinfo{author}{Bose, A.J.}, \bibinfo{author}{Jain, A.},
  \bibinfo{author}{Molino, P.}, \bibinfo{author}{Hamilton, W.L.},
  \bibinfo{year}{2019}.
\newblock \bibinfo{title}{Meta-graph: Few shot link prediction via meta
  learning}.
\newblock \bibinfo{journal}{arXiv preprint arXiv:1912.09867} .
\bibitem[{Cavallari et~al.(2019)Cavallari, Cambria, Cai, Chang and
  Zheng}]{cavallari2019embedding}
\bibinfo{author}{Cavallari, S.}, \bibinfo{author}{Cambria, E.},
  \bibinfo{author}{Cai, H.}, \bibinfo{author}{Chang, K.C.C.},
  \bibinfo{author}{Zheng, V.W.}, \bibinfo{year}{2019}.
\newblock \bibinfo{title}{Embedding both finite and infinite communities on
  graphs [application notes]}.
\newblock \bibinfo{journal}{IEEE Computational Intelligence Magazine}
  \bibinfo{volume}{14}, \bibinfo{pages}{39--50}.
\bibitem[{Chen et~al.(2019)Chen, Zhang, Zhang, Chen and Chen}]{chen2019meta}
\bibinfo{author}{Chen, M.}, \bibinfo{author}{Zhang, W.},
  \bibinfo{author}{Zhang, W.}, \bibinfo{author}{Chen, Q.},
  \bibinfo{author}{Chen, H.}, \bibinfo{year}{2019}.
\newblock \bibinfo{title}{Meta relational learning for few-shot link prediction
  in knowledge graphs}, in: \bibinfo{booktitle}{Proceedings of the 2019
  Conference on Empirical Methods in Natural Language Processing and the 9th
  International Joint Conference on Natural Language Processing
  (EMNLP-IJCNLP)}, pp. \bibinfo{pages}{4208--4217}.
\bibitem[{Chen et~al.(2020)Chen, Jia and Xiang}]{chen2020review}
\bibinfo{author}{Chen, X.}, \bibinfo{author}{Jia, S.}, \bibinfo{author}{Xiang,
  Y.}, \bibinfo{year}{2020}.
\newblock \bibinfo{title}{A review: Knowledge reasoning over knowledge graph}.
\newblock \bibinfo{journal}{Expert Systems with Applications}
  \bibinfo{volume}{141}, \bibinfo{pages}{112948}.
\bibitem[{Dess{\i} et~al.()Dess{\i}, Osborne, Recupero, Buscaldi, Motta and
  Sack}]{dessi2020ai}
\bibinfo{author}{Dess{\i}, D.}, \bibinfo{author}{Osborne, F.},
  \bibinfo{author}{Recupero, D.R.}, \bibinfo{author}{Buscaldi, D.},
  \bibinfo{author}{Motta, E.}, \bibinfo{author}{Sack, H.}, .
\newblock \bibinfo{title}{Ai-kg: an automatically generated knowledge graph of
  artificial intelligence} .
\bibitem[{F{\"a}rber(2019)}]{farber2019microsoft}
\bibinfo{author}{F{\"a}rber, M.}, \bibinfo{year}{2019}.
\newblock \bibinfo{title}{The microsoft academic knowledge graph: A linked data
  source with 8 billion triples of scholarly data}, in:
  \bibinfo{booktitle}{International Semantic Web Conference},
  \bibinfo{organization}{Springer}. pp. \bibinfo{pages}{113--129}.
\bibitem[{Gong et~al.(2019)Gong, Gao, Liang, Shen, Wang and
  Lin}]{gong2019graphonomy}
\bibinfo{author}{Gong, K.}, \bibinfo{author}{Gao, Y.}, \bibinfo{author}{Liang,
  X.}, \bibinfo{author}{Shen, X.}, \bibinfo{author}{Wang, M.},
  \bibinfo{author}{Lin, L.}, \bibinfo{year}{2019}.
\newblock \bibinfo{title}{Graphonomy: Universal human parsing via graph
  transfer learning}, in: \bibinfo{booktitle}{Proceedings of the IEEE
  Conference on Computer Vision and Pattern Recognition}, pp.
  \bibinfo{pages}{7450--7459}.
\bibitem[{Goyal and Ferrara(2018)}]{goyal2018graph}
\bibinfo{author}{Goyal, P.}, \bibinfo{author}{Ferrara, E.},
  \bibinfo{year}{2018}.
\newblock \bibinfo{title}{Graph embedding techniques, applications, and
  performance: A survey}.
\newblock \bibinfo{journal}{Knowledge-Based Systems} \bibinfo{volume}{151},
  \bibinfo{pages}{78--94}.
\bibitem[{Groth et~al.(2010)Groth, Gibson and Velterop}]{groth2010ana}
\bibinfo{author}{Groth, P.}, \bibinfo{author}{Gibson, A.},
  \bibinfo{author}{Velterop, J.}, \bibinfo{year}{2010}.
\newblock \bibinfo{title}{The anatomy of a nanopublication}.
\newblock \bibinfo{journal}{Information Services \& Use} \bibinfo{volume}{30},
  \bibinfo{pages}{51--56}.
\bibitem[{Henk et~al.(2019)Henk, Vahdati, Nayyeri, Ali, Yazdi and
  Lehmann}]{henck2019}
\bibinfo{author}{Henk, V.}, \bibinfo{author}{Vahdati, S.},
  \bibinfo{author}{Nayyeri, M.}, \bibinfo{author}{Ali, M.},
  \bibinfo{author}{Yazdi, H.S.}, \bibinfo{author}{Lehmann, J.},
  \bibinfo{year}{2019}.
\newblock \bibinfo{title}{Metaresearch recommendations using knowledge graph
  embeddings}, in: \bibinfo{booktitle}{RecNLP workshop of AAAI Conference}.
\bibitem[{Jaradeh et~al.(2019)Jaradeh, Oelen, Farfar, Prinz, D'Souza,
  Kismih{\'o}k, Stocker and Auer}]{jaradeh2019open}
\bibinfo{author}{Jaradeh, M.Y.}, \bibinfo{author}{Oelen, A.},
  \bibinfo{author}{Farfar, K.E.}, \bibinfo{author}{Prinz, M.},
  \bibinfo{author}{D'Souza, J.}, \bibinfo{author}{Kismih{\'o}k, G.},
  \bibinfo{author}{Stocker, M.}, \bibinfo{author}{Auer, S.},
  \bibinfo{year}{2019}.
\newblock \bibinfo{title}{Open research knowledge graph: Next generation
  infrastructure for semantic scholarly knowledge}, in:
  \bibinfo{booktitle}{Proceedings of the 10th International Conference on
  Knowledge Capture}, pp. \bibinfo{pages}{243--246}.
\bibitem[{Ji et~al.(2021)Ji, Pan, Cambria, Marttinen and Yu}]{ji2020survey}
\bibinfo{author}{Ji, S.}, \bibinfo{author}{Pan, S.}, \bibinfo{author}{Cambria,
  E.}, \bibinfo{author}{Marttinen, P.}, \bibinfo{author}{Yu, P.S.},
  \bibinfo{year}{2021}.
\newblock \bibinfo{title}{A survey on knowledge graphs: Representation,
  acquisition and applications}.
\newblock \bibinfo{journal}{IEEE Transactions on Neural Networks and Learning
  Systems} \bibinfo{volume}{32}.
\bibitem[{Knoth and Zdrahal(2011)}]{core2011}
\bibinfo{author}{Knoth, P.}, \bibinfo{author}{Zdrahal, Z.},
  \bibinfo{year}{2011}.
\newblock \bibinfo{title}{Core: connecting repositories in the open access
  domain}, in: \bibinfo{booktitle}{CERN Workshop on Innovations in Scholarly
  Communication (OAI7)}.
\newblock \URLprefix \url{http://oro.open.ac.uk/32560/}. \bibinfo{note}{poster
  Session ID: 53}.
\bibitem[{Knoth and Zdrahal(2012)}]{knoth2012core}
\bibinfo{author}{Knoth, P.}, \bibinfo{author}{Zdrahal, Z.},
  \bibinfo{year}{2012}.
\newblock \bibinfo{title}{Core: three access levels to underpin open access}.
\newblock \bibinfo{journal}{D-Lib Magazine} \bibinfo{volume}{18},
  \bibinfo{pages}{1--13}.
\bibitem[{Kuhn et~al.(2016)Kuhn, Chichester, Krauthammer, Queralt-Rosinach,
  Verborgh, Giannakopoulos, Ngomo, Viglianti and
  Dumontier}]{kuhn2016decentralized}
\bibinfo{author}{Kuhn, T.}, \bibinfo{author}{Chichester, C.},
  \bibinfo{author}{Krauthammer, M.}, \bibinfo{author}{Queralt-Rosinach, N.},
  \bibinfo{author}{Verborgh, R.}, \bibinfo{author}{Giannakopoulos, G.},
  \bibinfo{author}{Ngomo, A.C.N.}, \bibinfo{author}{Viglianti, R.},
  \bibinfo{author}{Dumontier, M.}, \bibinfo{year}{2016}.
\newblock \bibinfo{title}{Decentralized provenance-aware publishing with
  nanopublications}.
\newblock \bibinfo{journal}{PeerJ Computer Science} \bibinfo{volume}{2},
  \bibinfo{pages}{e78}.
\bibitem[{Lacroix et~al.(2018)Lacroix, Usunier and
  Obozinski}]{lacroix2018canonical}
\bibinfo{author}{Lacroix, T.}, \bibinfo{author}{Usunier, N.},
  \bibinfo{author}{Obozinski, G.}, \bibinfo{year}{2018}.
\newblock \bibinfo{title}{Canonical tensor decomposition for knowledge base
  completion}, in: \bibinfo{booktitle}{International Conference on Machine
  Learning}, pp. \bibinfo{pages}{2863--2872}.
\bibitem[{Lee et~al.(2017)Lee, Kim, Lee and Yoon}]{lee2017transfer}
\bibinfo{author}{Lee, J.}, \bibinfo{author}{Kim, H.}, \bibinfo{author}{Lee,
  J.}, \bibinfo{author}{Yoon, S.}, \bibinfo{year}{2017}.
\newblock \bibinfo{title}{Transfer learning for deep learning on
  graph-structured data.}, in: \bibinfo{booktitle}{AAAI}, pp.
  \bibinfo{pages}{2154--2160}.
\bibitem[{Li et~al.(2019)Li, Wang, Wang, Jiang, Tang, Yan, Wang and
  Liu}]{li2019prtransh}
\bibinfo{author}{Li, L.}, \bibinfo{author}{Wang, P.}, \bibinfo{author}{Wang,
  Y.}, \bibinfo{author}{Jiang, J.}, \bibinfo{author}{Tang, B.},
  \bibinfo{author}{Yan, J.}, \bibinfo{author}{Wang, S.}, \bibinfo{author}{Liu,
  Y.}, \bibinfo{year}{2019}.
\newblock \bibinfo{title}{Prtransh: Embedding probabilistic medical knowledge
  from real world emr data}.
\newblock \bibinfo{journal}{arXiv preprint arXiv:1909.00672} .
\bibitem[{Maaten and Hinton(2008)}]{maaten2008visualizing}
\bibinfo{author}{Maaten, L.v.d.}, \bibinfo{author}{Hinton, G.},
  \bibinfo{year}{2008}.
\newblock \bibinfo{title}{Visualizing data using t-sne}.
\newblock \bibinfo{journal}{Journal of machine learning research}
  \bibinfo{volume}{9}, \bibinfo{pages}{2579--2605}.
\bibitem[{Mannocci et~al.(2019)Mannocci, Osborne and
  Motta}]{mannocci2019geographical}
\bibinfo{author}{Mannocci, A.}, \bibinfo{author}{Osborne, F.},
  \bibinfo{author}{Motta, E.}, \bibinfo{year}{2019}.
\newblock \bibinfo{title}{Geographical trends in academic conferences: An
  analysis of authors’ affiliations}.
\newblock \bibinfo{journal}{Data Science} \bibinfo{volume}{2},
  \bibinfo{pages}{181--203}.
\bibitem[{Nathani et~al.(2019)Nathani, Chauhan, Sharma and
  Kaul}]{nathani2019learning}
\bibinfo{author}{Nathani, D.}, \bibinfo{author}{Chauhan, J.},
  \bibinfo{author}{Sharma, C.}, \bibinfo{author}{Kaul, M.},
  \bibinfo{year}{2019}.
\newblock \bibinfo{title}{Learning attention-based embeddings for relation
  prediction in knowledge graphs}, in: \bibinfo{booktitle}{Proceedings of the
  57th Annual Meeting of the Association for Computational Linguistics}, pp.
  \bibinfo{pages}{4710--4723}.
\bibitem[{Nayyeri et~al.(2020a)Nayyeri, Vahdati, Zhou, Yazdi and
  Lehmann}]{nayyeri2020embedding}
\bibinfo{author}{Nayyeri, M.}, \bibinfo{author}{Vahdati, S.},
  \bibinfo{author}{Zhou, X.}, \bibinfo{author}{Yazdi, H.S.},
  \bibinfo{author}{Lehmann, J.}, \bibinfo{year}{2020}a.
\newblock \bibinfo{title}{Embedding-based recommendations on scholarly
  knowledge graphs}, in: \bibinfo{booktitle}{European Semantic Web Conference},
  \bibinfo{organization}{Springer}. pp. \bibinfo{pages}{255--270}.
\bibitem[{Nayyeri et~al.(2020b)Nayyeri, Xu, Vahdati, Vassilyeva, Sallinger,
  Yazdi and Lehmann}]{nayyeri2020fantastic}
\bibinfo{author}{Nayyeri, M.}, \bibinfo{author}{Xu, C.},
  \bibinfo{author}{Vahdati, S.}, \bibinfo{author}{Vassilyeva, N.},
  \bibinfo{author}{Sallinger, E.}, \bibinfo{author}{Yazdi, H.S.},
  \bibinfo{author}{Lehmann, J.}, \bibinfo{year}{2020}b.
\newblock \bibinfo{title}{Fantastic knowledge graph embeddings and how to find
  the right space for them}, in: \bibinfo{booktitle}{International Semantic Web
  Conference}, \bibinfo{organization}{Springer}. pp. \bibinfo{pages}{438--455}.
\bibitem[{Nayyeri et~al.(2020c)Nayyeri, Xu, Vahdati, Vassilyeva, Sallinger,
  Yazdi and Lehmann}]{iswcfantastic}
\bibinfo{author}{Nayyeri, M.}, \bibinfo{author}{Xu, C.},
  \bibinfo{author}{Vahdati, S.}, \bibinfo{author}{Vassilyeva, N.},
  \bibinfo{author}{Sallinger, E.}, \bibinfo{author}{Yazdi, H.S.},
  \bibinfo{author}{Lehmann, J.}, \bibinfo{year}{2020}c.
\newblock \bibinfo{title}{Fantastic knowledge graph embeddings and how to find
  the right space for them}, in: \bibinfo{booktitle}{ISWC}.
\bibitem[{Nickel et~al.(2016)Nickel, Murphy, Tresp and
  Gabrilovich}]{Nickel2015ReviewRelationalMLforKG}
\bibinfo{author}{Nickel, M.}, \bibinfo{author}{Murphy, K.},
  \bibinfo{author}{Tresp, V.}, \bibinfo{author}{Gabrilovich, E.},
  \bibinfo{year}{2016}.
\newblock \bibinfo{title}{A review of relational machine learning for knowledge
  graphs}.
\newblock \bibinfo{journal}{Proceedings of the IEEE} \bibinfo{volume}{104}.
\bibitem[{Nuzzolese et~al.(2016)Nuzzolese, Gentile, Presutti and
  Gangemi}]{nuzzolese2016semantic}
\bibinfo{author}{Nuzzolese, A.G.}, \bibinfo{author}{Gentile, A.L.},
  \bibinfo{author}{Presutti, V.}, \bibinfo{author}{Gangemi, A.},
  \bibinfo{year}{2016}.
\newblock \bibinfo{title}{Semantic web conference ontology-a refactoring
  solution}, in: \bibinfo{booktitle}{European Semantic Web Conference},
  \bibinfo{organization}{Springer}. pp. \bibinfo{pages}{84--87}.
\bibitem[{Osborne et~al.(2017)Osborne, Mannocci and Motta}]{osborne2017}
\bibinfo{author}{Osborne, F.}, \bibinfo{author}{Mannocci, A.},
  \bibinfo{author}{Motta, E.}, \bibinfo{year}{2017}.
\newblock \bibinfo{title}{Forecasting the spreading of technologies in research
  communities}, in: \bibinfo{booktitle}{Proceedings of the Knowledge Capture
  Conference}, \bibinfo{publisher}{ACM}, \bibinfo{address}{New York, NY, USA}.
  pp. \bibinfo{pages}{1:1--1:8}.
\newblock \DOIprefix\doi{10.1145/3148011.3148030}.
\bibitem[{Paliwal et~al.(2019)Paliwal, Gimeno, Nair, Li, Lubin, Kohli and
  Vinyals}]{paliwal2019regal}
\bibinfo{author}{Paliwal, A.}, \bibinfo{author}{Gimeno, F.},
  \bibinfo{author}{Nair, V.}, \bibinfo{author}{Li, Y.}, \bibinfo{author}{Lubin,
  M.}, \bibinfo{author}{Kohli, P.}, \bibinfo{author}{Vinyals, O.},
  \bibinfo{year}{2019}.
\newblock \bibinfo{title}{Regal: Transfer learning for fast optimization of
  computation graphs}.
\newblock \bibinfo{journal}{arXiv preprint arXiv:1905.02494} .
\bibitem[{Peroni and Shotton(2018)}]{peroni2018spar}
\bibinfo{author}{Peroni, S.}, \bibinfo{author}{Shotton, D.},
  \bibinfo{year}{2018}.
\newblock \bibinfo{title}{The spar ontologies}, in:
  \bibinfo{booktitle}{International Semantic Web Conference},
  \bibinfo{organization}{Springer}. pp. \bibinfo{pages}{119--136}.
\bibitem[{Peroni and Shotton(2020)}]{peroni2020opencitations}
\bibinfo{author}{Peroni, S.}, \bibinfo{author}{Shotton, D.},
  \bibinfo{year}{2020}.
\newblock \bibinfo{title}{Opencitations, an infrastructure organization for
  open scholarship}.
\newblock \bibinfo{journal}{Quantitative Science Studies} \bibinfo{volume}{1},
  \bibinfo{pages}{428--444}.
\bibitem[{Salatino et~al.(2020a)Salatino, Osborne and
  Motta}]{salatino2020research}
\bibinfo{author}{Salatino, A.}, \bibinfo{author}{Osborne, F.},
  \bibinfo{author}{Motta, E.}, \bibinfo{year}{2020}a.
\newblock \bibinfo{title}{Researchflow: Understanding the knowledge flow
  between academia and industry}, in: \bibinfo{booktitle}{Knowledge Engineering
  and Knowledge Management – 22nd International Conference, EKAW 2020}.
\bibitem[{Salatino et~al.(2020b)Salatino, Thanapalasingam, Mannocci, Birukou,
  Osborne and Motta}]{salatino2020computer}
\bibinfo{author}{Salatino, A.A.}, \bibinfo{author}{Thanapalasingam, T.},
  \bibinfo{author}{Mannocci, A.}, \bibinfo{author}{Birukou, A.},
  \bibinfo{author}{Osborne, F.}, \bibinfo{author}{Motta, E.},
  \bibinfo{year}{2020}b.
\newblock \bibinfo{title}{The computer science ontology: A comprehensive
  automatically-generated taxonomy of research areas}.
\newblock \bibinfo{journal}{Data Intelligence} \bibinfo{volume}{2},
  \bibinfo{pages}{379--416}.
\bibitem[{Salatino et~al.(2018)Salatino, Thanapalasingam, Mannocci, Osborne and
  Motta}]{salatino2018computer}
\bibinfo{author}{Salatino, A.A.}, \bibinfo{author}{Thanapalasingam, T.},
  \bibinfo{author}{Mannocci, A.}, \bibinfo{author}{Osborne, F.},
  \bibinfo{author}{Motta, E.}, \bibinfo{year}{2018}.
\newblock \bibinfo{title}{The computer science ontology: a large-scale taxonomy
  of research areas}, in: \bibinfo{booktitle}{ISWC}, pp.
  \bibinfo{pages}{187--205}.
\bibitem[{Schneider et~al.(2014)Schneider, Ciccarese, Clark and
  Boyce}]{schneider2014using}
\bibinfo{author}{Schneider, J.}, \bibinfo{author}{Ciccarese, P.},
  \bibinfo{author}{Clark, T.}, \bibinfo{author}{Boyce, R.D.},
  \bibinfo{year}{2014}.
\newblock \bibinfo{title}{Using the micropublications ontology and the open
  annotation data model to represent evidence within a drug-drug interaction
  knowledge base}.
\bibitem[{Shotton(2009)}]{shotton2009semantic}
\bibinfo{author}{Shotton, D.}, \bibinfo{year}{2009}.
\newblock \bibinfo{title}{Semantic publishing: the coming revolution in
  scientific journal publishing}.
\newblock \bibinfo{journal}{Learned Publishing} \bibinfo{volume}{22},
  \bibinfo{pages}{85--94}.
\bibitem[{Stanovsky et~al.(2017)Stanovsky, Gruhl and
  Mendes}]{stanovsky2017recognizing}
\bibinfo{author}{Stanovsky, G.}, \bibinfo{author}{Gruhl, D.},
  \bibinfo{author}{Mendes, P.}, \bibinfo{year}{2017}.
\newblock \bibinfo{title}{Recognizing mentions of adverse drug reaction in
  social media using knowledge-infused recurrent models}, in:
  \bibinfo{booktitle}{Proceedings of the 15th Conference of the European
  Chapter of the Association for Computational Linguistics: Volume 1, Long
  Papers}, pp. \bibinfo{pages}{142--151}.
\bibitem[{Sun et~al.(2019a)Sun, Qu, Hoffmann, Huang and Tang}]{sun2019vgraph}
\bibinfo{author}{Sun, F.Y.}, \bibinfo{author}{Qu, M.},
  \bibinfo{author}{Hoffmann, J.}, \bibinfo{author}{Huang, C.W.},
  \bibinfo{author}{Tang, J.}, \bibinfo{year}{2019}a.
\newblock \bibinfo{title}{vgraph: A generative model for joint community
  detection and node representation learning}, in: \bibinfo{booktitle}{Advances
  in Neural Information Processing Systems}, pp. \bibinfo{pages}{514--524}.
\bibitem[{Sun et~al.(2019b)Sun, Deng, Nie and Tang}]{sun2018rotate}
\bibinfo{author}{Sun, Z.}, \bibinfo{author}{Deng, Z.H.}, \bibinfo{author}{Nie,
  J.Y.}, \bibinfo{author}{Tang, J.}, \bibinfo{year}{2019}b.
\newblock \bibinfo{title}{Rotate: Knowledge graph embedding by relational
  rotation in complex space}, in: \bibinfo{booktitle}{International Conference
  on Learning Representations}.
\newblock \URLprefix \url{https://openreview.net/forum?id=HkgEQnRqYQ}.
\bibitem[{Sun et~al.(2019c)Sun, Vashishth, Sanyal, Talukdar and
  Yang}]{sun2019re}
\bibinfo{author}{Sun, Z.}, \bibinfo{author}{Vashishth, S.},
  \bibinfo{author}{Sanyal, S.}, \bibinfo{author}{Talukdar, P.},
  \bibinfo{author}{Yang, Y.}, \bibinfo{year}{2019}c.
\newblock \bibinfo{title}{A re-evaluation of knowledge graph completion
  methods}.
\newblock \bibinfo{journal}{arXiv preprint arXiv:1911.03903} .
\bibitem[{Tran and Takasu(2019)}]{tran2019exploring}
\bibinfo{author}{Tran, H.N.}, \bibinfo{author}{Takasu, A.},
  \bibinfo{year}{2019}.
\newblock \bibinfo{title}{Exploring scholarly data by semantic query on
  knowledge graph embedding space}, in: \bibinfo{booktitle}{International
  Conference on Theory and Practice of Digital Libraries},
  \bibinfo{organization}{Springer}. pp. \bibinfo{pages}{154--162}.
\bibitem[{Trouillon et~al.(2016)Trouillon, Welbl, Riedel, Gaussier and
  Bouchard}]{trouillon2016complex}
\bibinfo{author}{Trouillon, T.}, \bibinfo{author}{Welbl, J.},
  \bibinfo{author}{Riedel, S.}, \bibinfo{author}{Gaussier, {\'E}.},
  \bibinfo{author}{Bouchard, G.}, \bibinfo{year}{2016}.
\newblock \bibinfo{title}{Complex embeddings for simple link prediction}, in:
  \bibinfo{booktitle}{International Conference on Machine Learning}, pp.
  \bibinfo{pages}{2071--2080}.
\bibitem[{Vu et~al.(2019)Vu, Nguyen, Nguyen, Phung et~al.}]{vu2019capsule}
\bibinfo{author}{Vu, T.}, \bibinfo{author}{Nguyen, T.D.},
  \bibinfo{author}{Nguyen, D.Q.}, \bibinfo{author}{Phung, D.}, et~al.,
  \bibinfo{year}{2019}.
\newblock \bibinfo{title}{A capsule network-based embedding model for knowledge
  graph completion and search personalization}, in:
  \bibinfo{booktitle}{Proceedings of the 2019 Conference of the North American
  Chapter of the Association for Computational Linguistics: Human Language
  Technologies, Volume 1 (Long and Short Papers)}, pp.
  \bibinfo{pages}{2180--2189}.
\bibitem[{Wang et~al.(2019a)Wang, Zhang, Zhang, Leskovec, Zhao, Li and
  Wang}]{wang2019knowledge}
\bibinfo{author}{Wang, H.}, \bibinfo{author}{Zhang, F.},
  \bibinfo{author}{Zhang, M.}, \bibinfo{author}{Leskovec, J.},
  \bibinfo{author}{Zhao, M.}, \bibinfo{author}{Li, W.}, \bibinfo{author}{Wang,
  Z.}, \bibinfo{year}{2019}a.
\newblock \bibinfo{title}{Knowledge-aware graph neural networks with label
  smoothness regularization for recommender systems}, in:
  \bibinfo{booktitle}{Proceedings of the 25th ACM SIGKDD International
  Conference on Knowledge Discovery \& Data Mining}, pp.
  \bibinfo{pages}{968--977}.
\bibitem[{Wang et~al.(2020a)Wang, Shen, Huang, Wu, Dong and
  Kanakia}]{wang2020microsoft}
\bibinfo{author}{Wang, K.}, \bibinfo{author}{Shen, Z.}, \bibinfo{author}{Huang,
  C.}, \bibinfo{author}{Wu, C.H.}, \bibinfo{author}{Dong, Y.},
  \bibinfo{author}{Kanakia, A.}, \bibinfo{year}{2020}a.
\newblock \bibinfo{title}{Microsoft academic graph: When experts are not
  enough}.
\newblock \bibinfo{journal}{Quantitative Science Studies} \bibinfo{volume}{1},
  \bibinfo{pages}{396--413}.
\bibitem[{Wang et~al.(2017)Wang, Mao, Wang and Guo}]{wang2017knowledge}
\bibinfo{author}{Wang, Q.}, \bibinfo{author}{Mao, Z.}, \bibinfo{author}{Wang,
  B.}, \bibinfo{author}{Guo, L.}, \bibinfo{year}{2017}.
\newblock \bibinfo{title}{Knowledge graph embedding: A survey of approaches and
  applications}.
\newblock \bibinfo{journal}{IEEE TKDE} \bibinfo{volume}{29}.
\bibitem[{Wang et~al.(2020b)Wang, Liu, Tang, Tuarob, Xia, Gong and
  King}]{wang2020attributed}
\bibinfo{author}{Wang, W.}, \bibinfo{author}{Liu, J.}, \bibinfo{author}{Tang,
  T.}, \bibinfo{author}{Tuarob, S.}, \bibinfo{author}{Xia, F.},
  \bibinfo{author}{Gong, Z.}, \bibinfo{author}{King, I.},
  \bibinfo{year}{2020}b.
\newblock \bibinfo{title}{Attributed collaboration network embedding for
  academic relationship mining}.
\newblock \bibinfo{journal}{ACM Transactions on the Web (TWEB)}
  \bibinfo{volume}{15}, \bibinfo{pages}{1--20}.
\bibitem[{Wang et~al.(2020c)Wang, Yao, Kwok and Ni}]{wang2020generalizing}
\bibinfo{author}{Wang, Y.}, \bibinfo{author}{Yao, Q.}, \bibinfo{author}{Kwok,
  J.T.}, \bibinfo{author}{Ni, L.M.}, \bibinfo{year}{2020}c.
\newblock \bibinfo{title}{Generalizing from a few examples: A survey on
  few-shot learning}.
\newblock \bibinfo{journal}{ACM Computing Surveys (CSUR)} \bibinfo{volume}{53},
  \bibinfo{pages}{1--34}.
\bibitem[{Wang et~al.(2019b)Wang, Ren, He, Zhang and Hu}]{wang2019robust}
\bibinfo{author}{Wang, Z.}, \bibinfo{author}{Ren, Z.}, \bibinfo{author}{He,
  C.}, \bibinfo{author}{Zhang, P.}, \bibinfo{author}{Hu, Y.},
  \bibinfo{year}{2019}b.
\newblock \bibinfo{title}{Robust embedding with multi-level structures for link
  prediction.}, in: \bibinfo{booktitle}{IJCAI}, pp.
  \bibinfo{pages}{5240--5246}.
\bibitem[{Wolstencroft et~al.(2013)Wolstencroft, Haines, Fellows, Williams,
  Withers, Owen et~al.}]{wolstencroft2013taverna}
\bibinfo{author}{Wolstencroft, K.}, \bibinfo{author}{Haines, R.},
  \bibinfo{author}{Fellows, D.}, \bibinfo{author}{Williams, A.},
  \bibinfo{author}{Withers, D.}, \bibinfo{author}{Owen, S.}, et~al.,
  \bibinfo{year}{2013}.
\newblock \bibinfo{title}{The taverna workflow suite: designing and executing
  workflows of web services on the desktop, web or in the cloud}.
\newblock \bibinfo{journal}{Nucleic acids research} \bibinfo{volume}{41},
  \bibinfo{pages}{W557--W561}.
\bibitem[{Wu et~al.(2020)Wu, Pan, Chen, Long, Zhang and
  Philip}]{wu2020comprehensive}
\bibinfo{author}{Wu, Z.}, \bibinfo{author}{Pan, S.}, \bibinfo{author}{Chen,
  F.}, \bibinfo{author}{Long, G.}, \bibinfo{author}{Zhang, C.},
  \bibinfo{author}{Philip, S.Y.}, \bibinfo{year}{2020}.
\newblock \bibinfo{title}{A comprehensive survey on graph neural networks}.
\newblock \bibinfo{journal}{IEEE Transactions on Neural Networks and Learning
  Systems} .
\bibitem[{Yao et~al.(2017)Yao, Zhang, Wei, Jin, Zhang, Zhang and
  Chen}]{yao2017incorporating}
\bibinfo{author}{Yao, L.}, \bibinfo{author}{Zhang, Y.}, \bibinfo{author}{Wei,
  B.}, \bibinfo{author}{Jin, Z.}, \bibinfo{author}{Zhang, R.},
  \bibinfo{author}{Zhang, Y.}, \bibinfo{author}{Chen, Q.},
  \bibinfo{year}{2017}.
\newblock \bibinfo{title}{Incorporating knowledge graph embeddings into topic
  modeling}, in: \bibinfo{booktitle}{Thirty-First AAAI Conference on Artificial
  Intelligence}.
\bibitem[{Yin(2020)}]{yin2020meta}
\bibinfo{author}{Yin, W.}, \bibinfo{year}{2020}.
\newblock \bibinfo{title}{Meta-learning for few-shot natural language
  processing: A survey}.
\newblock \bibinfo{journal}{arXiv preprint arXiv:2007.09604} .
\bibitem[{Zhang et~al.(2020)Zhang, Yao, Huang, Jiang, Li and
  Chawla}]{zhang2020few}
\bibinfo{author}{Zhang, C.}, \bibinfo{author}{Yao, H.}, \bibinfo{author}{Huang,
  C.}, \bibinfo{author}{Jiang, M.}, \bibinfo{author}{Li, Z.},
  \bibinfo{author}{Chawla, N.V.}, \bibinfo{year}{2020}.
\newblock \bibinfo{title}{Few-shot knowledge graph completion}, in:
  \bibinfo{booktitle}{Proceedings of the AAAI Conference on Artificial
  Intelligence}, pp. \bibinfo{pages}{3041--3048}.
\bibitem[{Zhang et~al.(2019)Zhang, Tay, Yao and Liu}]{quate2019zhang}
\bibinfo{author}{Zhang, S.}, \bibinfo{author}{Tay, Y.}, \bibinfo{author}{Yao,
  L.}, \bibinfo{author}{Liu, Q.}, \bibinfo{year}{2019}.
\newblock \bibinfo{title}{Quaternion knowledge graph embedding}.
\newblock \bibinfo{journal}{arXiv preprint arXiv:1904.10281} .
\bibitem[{Zhang et~al.(2018)Zhang, Zhang, Yao and Tang}]{zhang2018name}
\bibinfo{author}{Zhang, Y.}, \bibinfo{author}{Zhang, F.}, \bibinfo{author}{Yao,
  P.}, \bibinfo{author}{Tang, J.}, \bibinfo{year}{2018}.
\newblock \bibinfo{title}{Name disambiguation in aminer: Clustering,
  maintenance, and human in the loop.}, in: \bibinfo{booktitle}{Proceedings of
  the 24th ACM SIGKDD International Conference on Knowledge Discovery \& Data
  Mining}, pp. \bibinfo{pages}{1002--1011}.
\bibitem[{Zhou et~al.(2018)Zhou, Cui, Zhang, Yang, Liu, Wang, Li and
  Sun}]{zhou2018graph}
\bibinfo{author}{Zhou, J.}, \bibinfo{author}{Cui, G.}, \bibinfo{author}{Zhang,
  Z.}, \bibinfo{author}{Yang, C.}, \bibinfo{author}{Liu, Z.},
  \bibinfo{author}{Wang, L.}, \bibinfo{author}{Li, C.}, \bibinfo{author}{Sun,
  M.}, \bibinfo{year}{2018}.
\newblock \bibinfo{title}{Graph neural networks: A review of methods and
  applications}, in: \bibinfo{booktitle}{CoRR}.
\bibitem[{Zhuang et~al.(2020)Zhuang, Qi, Duan, Xi, Zhu, Zhu, Xiong and
  He}]{zhuang2020comprehensive}
\bibinfo{author}{Zhuang, F.}, \bibinfo{author}{Qi, Z.}, \bibinfo{author}{Duan,
  K.}, \bibinfo{author}{Xi, D.}, \bibinfo{author}{Zhu, Y.},
  \bibinfo{author}{Zhu, H.}, \bibinfo{author}{Xiong, H.}, \bibinfo{author}{He,
  Q.}, \bibinfo{year}{2020}.
\newblock \bibinfo{title}{A comprehensive survey on transfer learning}.
\newblock \bibinfo{journal}{Proceedings of the IEEE} .

\end{thebibliography}


\end{document}